\documentclass[final]{cvpr}

\usepackage{times}
\usepackage{epsfig}
\usepackage{graphicx}
\usepackage{amsmath}
\usepackage{amssymb}

\usepackage{subcaption}
\usepackage{caption}
\usepackage{multirow}
\usepackage{tabularx}
\usepackage[symbol]{footmisc}
\usepackage{verbatim}

\makeatletter
\@namedef{ver@everyshi.sty}{}
\makeatother
\usepackage{tikz}
\usepackage{pgfplots}
\usepackage{pgfplots}\pgfplotsset{compat=1.9}

\newcommand*{\affaddr}[1]{#1} % No op here. Customize it for different styles.
\newcommand*{\affmark}[1][*]{\textsuperscript{#1}}
\newcommand*{\email}[1]{#1}

\usepackage[pagebackref=true,breaklinks=true,colorlinks,bookmarks=false]{hyperref}

\def\cvprPaperID{6983} % *** Enter the CVPR Paper ID here
\def\confYear{CVPR 2021}

\begin{document}

%%%%%%%%% TITLE
\title{GAN Prior Embedded Network for Blind Face Restoration in the Wild}

%\author{Tao Yang\\
% For a paper whose authors are all at the same institution,
% omit the following lines up until the closing ``}''.
% Additional authors and addresses can be added with ``\and'',
% just like the second author.
% To save space, use either the email address or home page, not both
%\and
%Peiran Ren\\
%DAMO Academy, Alibaba Group\\
%{\tt\small secondauthor@i2.org}
%}
%\author[1]{Tao Yang\thanks{yangtao9009@gmail.com}}
%\author[1]{Peiran Ren\thanks{peiran_r@sohu.com}}
%\author[1]{Xuansong Xie\thanks{xingtong.xxs@taobao.com}}
%\author[1,2]{Lei Zhang\thanks{cslzhang@comp.polyu.edu.hk}}
%\affil[1]{DAMO Academy, Alibaba Group}
%\affil[2]{Department of Computing, The Hong Kong Polytechnic University}

\author{%
Tao Yang\affmark[1], Peiran Ren\affmark[1], Xuansong Xie\affmark[1], and Lei Zhang\affmark[1,2]\footnotemark\\
\affaddr{\affmark[1]DAMO Academy, Alibaba Group}\\
\affaddr{\affmark[2]Department of Computing, The Hong Kong Polytechnic University}\\
\email{\tt\small{yangtao9009@gmail.com, peiran\_r@sohu.com, xingtong.xxs@taobao.com, cslzhang@comp.polyu.edu.hk}}\\
%\affaddr{\LaTeX\ University}%
}

\maketitle
%\twocolumn[{%
%\renewcommand\twocolumn[1][]{#1}%
%\maketitle
%\begin{center}
%    \centering
%    \includegraphics[width=\textwidth]{imgs/oldphoto-compressed.pdf}
%    \captionof{figure}{Restored group photo of the Solvay Conference, 1927. Our model can be applied to restore faces in old photos. Best viewed by zooming to 400\% plus in the screen.}
%\label{fig:oldphotos}
%\end{center}%
%}]
%\thispagestyle{empty}

%%%%%%%%% ABSTRACT
\begin{abstract}
   Blind face restoration (BFR) from severely degraded face images in the wild is a very challenging problem. Due to the high illness of the problem and the complex unknown degradation, directly training a deep neural network (DNN) usually cannot lead to acceptable results. Existing generative adversarial network (GAN) based methods can produce better results but tend to generate over-smoothed restorations. In this work, we propose a new method by first learning a GAN for high-quality face image generation and embedding it into a U-shaped DNN as a prior decoder, then fine-tuning the GAN prior embedded DNN with a set of synthesized low-quality face images. The GAN blocks are designed to ensure that the latent code and noise input to the GAN can be respectively generated from the deep and shallow features of the DNN, controlling the global face structure, local face details and background of the reconstructed image. The proposed GAN prior embedded network (GPEN) is easy-to-implement, and it can generate visually photo-realistic results. Our experiments demonstrated that the proposed GPEN achieves significantly superior results to state-of-the-art BFR methods both quantitatively and qualitatively, especially for the restoration of severely degraded face images in the wild. The source code and models can be found at \url{https://github.com/yangxy/GPEN}. 
\end{abstract}

%%%%%%%%% BODY TEXT
\section{Introduction}
\footnotetext[1]{This work is partially supported by the Hong Kong RGC RIF grant (R5001-18).}

Face images are among the most popular types of images in our daily life, while face images are often degraded due to the many factors such as low resolution, blur, noise, compression, etc., or the combination of them. Face image restoration has been attracting significant attentions, aiming at reproducing a clear and realistic face image from the degraded input. Traditional face image restoration methods \cite{Zhang2011Sparseprior,Bourlai2011Restore,Bakerand2000Hallucinate,Nishiyama2009Face} usually solve an inverse problem based on the degradation model and handcrafted priors, which demonstrate limited performance in practice. Recently, deep neural networks (DNNs) have shown superior results in a variety of computer vision tasks \cite{Kupyn2017DeblurGAN,Yu2018Deepfill,Guo2019CBDNet,Ledig2017SRGAN,Lim2017EDSR}, and many DNN based face restoration methods \cite{Yu2017Hallucinating,Li2018GFRNet,Hu20203dprior} have also been developed and they have demonstrated much better performance than traditional ones. 

Though much progress has been made for face restoration, blind face restoration (BFR) remains a challenging research problem because of the unknown and complex degradation of low quality (LQ) face images in the wild. In order to recover a high-quality (HQ) face image with photo-realistic textures from an LQ face image, a number of BFR methods have been proposed by resorting to the spatial transformer networks \cite{Yu2017Hallucinating}, exemplar images \cite{Li2018GFRNet,Li2020ASFFNet,Dogan2019Exemplar}, 3D facial priors \cite{Hu20203dprior}, and facial component dictionaries \cite{Li2020Restore}. Yang \emph{et al}. \cite{Yang2020HiFaceGANFR} proposed a collaborative suppression and replenishment (CSR) approach to progressively replenish facial details. These methods exhibit impressive results on artificially degraded faces; however, they fail to tackle real-world LQ face images. The conditional generative adversarial network (cGAN) based methods such as Pix2Pix \cite{Isola2017Pix2Pix} and Pix2PixHD \cite{Wang2018Pix2PixHD} learn a direct mapping from input image to output image. These methods achieve more realistic results but tend to over-smooth the images (see Figures~\ref{fig:comp} and \ref{fig:realcomp}), which is commonly blamed to the high illness of real-world BFR tasks. 

With the rapid advancement of GAN techniques \cite{Karras2018StyleGAN,Karras2019StyleGAN2}, recently some methods have been proposed to reconstruct faces from extremely low resolution inputs \cite{Gu2019Prior,Menon2020PULSE,Richardson2020pSp}. Richardson \emph{et al}. \cite{Richardson2020pSp} employed an encoder network to generate a series of style vectors before feeding them into a pre-trained generator, achieving a generic image-to-image translation framework. However, such methods can only work on non-blind image super-resolution problems. Furthermore, they kept the pre-trained GAN unchanged in training for the consistency and convenience of face manipulations. This however leads to unstable quality of restored faces when dealing with real-world LQ face images with complex background, because it is hard to accurately project a face image with limited resolution to a desired latent code (e.g., a vector of size $512$ in StyleGAN \cite{Karras2018StyleGAN,Karras2019StyleGAN2}).

\begin{figure}
   \includegraphics[width=.5\textwidth]{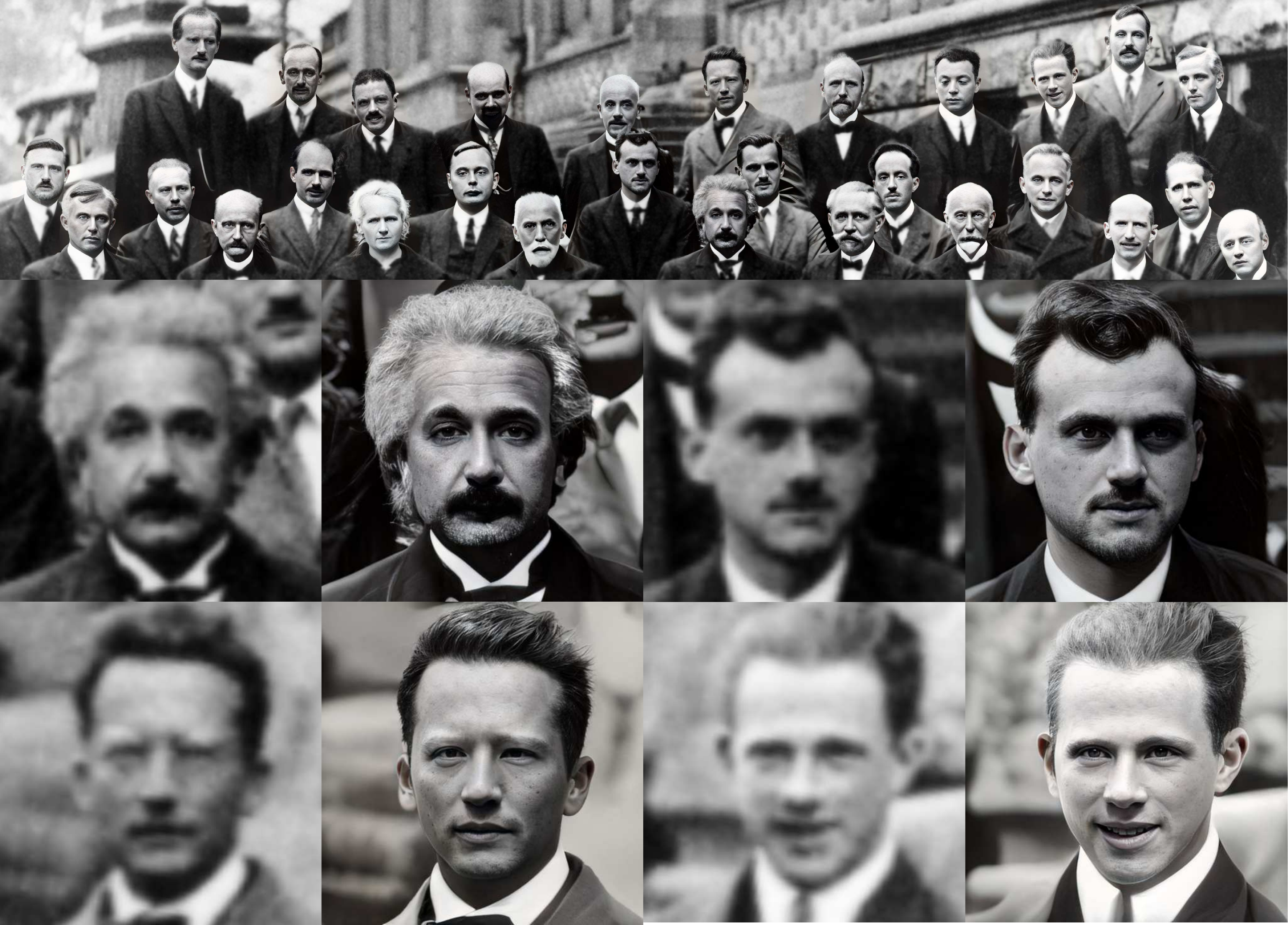}
   \caption{Restored face images from the group photo taken in the Solvay Conference, 1927. Best viewed by zooming to $200\%$ in the screen.}
\label{fig:oldphotos}
\vspace*{-5mm}
\end{figure}

In this work, we revisit the problem of BFR and target at restoring HQ faces from degraded face observations in the wild. Our idea is to seamlessly integrate the advantages of GAN and DNN. We first pre-train a GAN for HQ face image generation and embed it into a DNN as a decoder prior for face restoration. The GAN prior embedded DNN is then fine-tuned by a set of synthesized LQ-HQ face image pairs, during which the DNN learns to map the input degraded image to a desired latent space so that the GAN prior network can reproduce the desired HQ face images. We carefully design the GAN blocks to make them well suited for a U-shaped DNN, where the deep features are used to generate the latent code for global face reproduction, while the shallow features are used as noise to generate local face details and keep the image background. In this way, our learned model can reconstruct HQ faces with photo-realistic details from even severely degraded face images in the wild, avoiding over-smoothed results caused by the high illness of the BFR problem. Figure~\ref{fig:oldphotos} shows an example. One can see that our model reconstructs the face images of those great scientists with clear details from the old photo taken in 1927. 

\vspace*{2mm}
The main contributions of this work are summarized as follows:
\vspace*{-2mm}
\begin{itemize}
  \setlength{\itemsep}{0.7pt}
  \setlength{\parsep}{0.7pt}
  \setlength{\parskip}{0.7pt}
   %\item We propose to embed a GAN prior network into a DNN, and fine-tune it for effective BFR. Note that previous works can only transfer the pre-trained GAN into a network without fine-tuning.
   %\item We modify the GAN blocks so that they can be easily embedded into a U-shaped DNN for fine-tuning. The latent code and noise input of the GAN are respectively generated from the deep and shallow features of the DNN, reconstructing the global structure and local details of the face accordingly.
   %\item Our model sets new state-of-the-art in BFR. It is capable of tackling severely degraded faces in real-world scenarios.
   \item We learn and embed a GAN prior network into a DNN, and fine-tune the GAN embedded DNN for effective BFR in the wild. It is worthy to note that previous works only transfer the pre-trained GAN into a network without fine-tuning.
   \item The GAN blocks are designed so that they can be easily embedded into a U-shaped DNN for fine-tuning. The latent code and noise input of the GAN are respectively generated from the deep and shallow features of the DNN to reconstruct the global structure, local face details and background of the image accordingly.
   \item Our model sets new state-of-the-art in BFR. It is capable of tackling severely degraded face images taken in real-world scenarios.
\end{itemize}
\vspace*{-2mm}

\section{Related Work}
%Our method is obviously related to face image restoration and GAN based synthesis methods. In this section, we start by briefly discussing recent developments of face image restoration and generative adversarial networks, then providing an overview of the image prior closest to this paper.

\textbf{Face Image Restoration.}
As a specific but important branch of image restoration, face image restoration has been widely studied for many years. In the early stage, Zhang \emph{et al}. \cite{Zhang2011Sparseprior} presented a joint blind image restoration and recognition method by using sparse representation to handle face recognition from LQ images. Nishiyama \emph{et al}. \cite{Nishiyama2009Face} proposed to improve the recognition performance of blurry faces by using a pre-defined set of blur kernels to restore them. With the unprecedented success of DNNs in solving image restoration tasks such as denoising \cite{Guo2019CBDNet}, deblurring \cite{Kupyn2017DeblurGAN,Shen2018Deblur}, inpainting \cite{Yu2018Deepfill,Liu2018Partialconv} and image super-resolution \cite{Ledig2017SRGAN,Lim2017EDSR}, many DNN based face image restoration methods have also been proposed \cite{Chen2018FSRNet,Kim2019PFSR,Ma2020FSR}, which advance the traditional methods by a large margin. Considering the fact that facial images have specific structures, it is interesting to investigate whether we can restore a clear face image from severely degraded ones without knowing the degradation model. The so-called blind face restoration (BFR) problem has been attracting intensive research attentions in recent years \cite{Li2018GFRNet,Dogan2019Exemplar,Li2020Restore,Yang2020HiFaceGANFR}, while it is still a challenging task due to the complex image degradations in the wild.

Huang \emph{et al}. \cite{Huang2017Wavelet} presented a wavelet-based approach that can ultra-resolve a very low-resolution (LR) face image. Chen \emph{et al}. \cite{Chen2018FSRNet} learned the facial geometry prior to recover the high-resolution (HR) faces. Ma \emph{et al}. \cite{Ma2020FSR} performed face super-resolution with iterative collaboration between two recurrent networks on facial image recovery and landmark estimation, respectively. Li \emph{et al}. \cite{Li2018GFRNet} used a guiding image and a wrapper subnetwork to cope with appearance variations between the LR input and the HR guiding image. This work was further extended by using an unconstrained HR face image \cite{Dogan2019Exemplar}, multi-exemplar images \cite{Li2020ASFFNet}, and multi-scale component dictionaries \cite{Li2020Restore}. Hu \emph{et al}. \cite{Hu20203dprior} explicitly incorporated 3D facial priors to grasp the sharp facial structures. A collaborative suppression and replenishment approach was proposed by Yang \emph{et al}. \cite{Yang2020HiFaceGANFR} to progressively replenish facial details. Existing works have generated impressive results on artificially degraded faces, but often failed in real-world scenarios due to the complex unknown degradation. Furthermore, their performance depends heavily on the accurate facial prior knowledge which however is hard to obtain from severely degraded face images in the wild, leading to unpredictable failures.

\textbf{Generative Adversarial Network (GAN).}
Since the seminal work by Goodfellow \emph{et al}. \cite{Goodfellow2014GAN}, great progress has been accomplished on learning GAN models \cite{Karras2018PGGAN,Brock2019BigGAN,Karras2018StyleGAN,Karras2019StyleGAN2}. GAN has been widely used for various computer vision applications due to its powerful ability to generate photo-realistic images. Some typical applications include image inpainting \cite{Yu2018Deepfill}, super-resolution \cite{Ledig2017SRGAN,Wang2018ESRGAN}, image colorization \cite{Isola2017Pix2Pix,Suarez2017Color}, texture synthesis \cite{Slossberg2018Texture}, etc. Particularly, to provide more user controls for image synthesis, conditional GAN (cGAN) has been proposed \cite{Mirza2014cGAN}. By feeding the generator with different conditional information \cite{Park2019SPADE,Isola2017Pix2Pix,Zhu2017CycleGAN}, cGANs succeed in handling various image-to-image translation problems. Isola \emph{et al}. \cite{Isola2017Pix2Pix} showed that the conditional adversarial networks can be used as a general-purpose solution to image-to-image translation problems. Many following works, such as unsupervised learning \cite{Zhu2017CycleGAN}, disentangled learning \cite{Lee2018DRIT}, few-shot learning \cite{Liu2019Few}, high resolution image synthesis \cite{Wang2018Pix2PixHD}, multi-domain translation \cite{Choi2018Stargan}, multi-modal translation \cite{Zhu2017Multimodal}, have been proposed to extend cGAN to different scenarios. The cGAN learns a direct mapping from the input domain to the output one. Unfortunately, the generated results by cGANs are usually over-smoothed in highly ill-posed tasks such as BFR.

\textbf{GAN Prior for Image Generation.}
Deep generative models are popular in solving many inverse problems, e.g. deblurring \cite{Kupyn2017DeblurGAN}, image inpainting \cite{Yu2018Deepfill}, phase retrieval \cite{Hand2018Phase}, etc. Recently, many works have been developed for the task of GAN inversion, i.e., reversing a given image back to a latent code with a pre-trained GAN model. Existing methods either optimize the latent code \cite{Abdal2019Img2StyleGAN} or learn an extra encoder to project the image space back to the latent space \cite{Gu2019Prior,Richardson2020pSp}. Abdal \emph{et al}. \cite{Abdal2019Img2StyleGAN} embedded images into an extended latent space of StyleGAN, allowing further semantic image editing operations. Gu \emph{et al}. \cite{Gu2019Prior} employed multiple latent codes to generate multiple feature maps to output the final image. These optimization-based methods, however, are slow and improper for real-world applications. To address this issue, Pixel2Style2Pixel (pSp) \cite{Richardson2020pSp} embeds real images into extended latent space without additional optimization, which can be used in a wide range of image-to-image translation tasks. Menon \emph{et al}. \cite{Menon2020PULSE} proposed a self-supervised approach that traverses the HR natural image manifold, searching for images that can downscale to the original LR image. GAN inversion is an important step for applying GANs to real-world applications. However, it is difficult to perfectly project the image space back to the latent space. Moreover, it is hard, if not possible, to invert a blindly degraded face into a latent space.

Some works were proposed to transfer GAN priors. Wang \emph{et al}. \cite{Wang2018TransferGAN} applied domain adaptation to image generation with
GANs. They further proposed a novel knowledge transfer method for generative models by using a knowledge mining network \cite{Wang2020MineGAN}. Fr{\'e}gier and Gouray \cite{Frgier2019Mind2MindT} introduced a novel approach for transfer learning with GAN architecture. These works target at transferring the knowledge from the source domain to different target domains, while in our work, the source and target domains are the same. We embed the GAN prior learned for face generation into a DNN for face restoration, and jointly fine-tune the GAN prior network with the DNN so that the latent code and noise input can be well generated from the degraded face image at different network layers.

\begin{figure}[t!]
\centering
\includegraphics[width=0.48\textwidth]{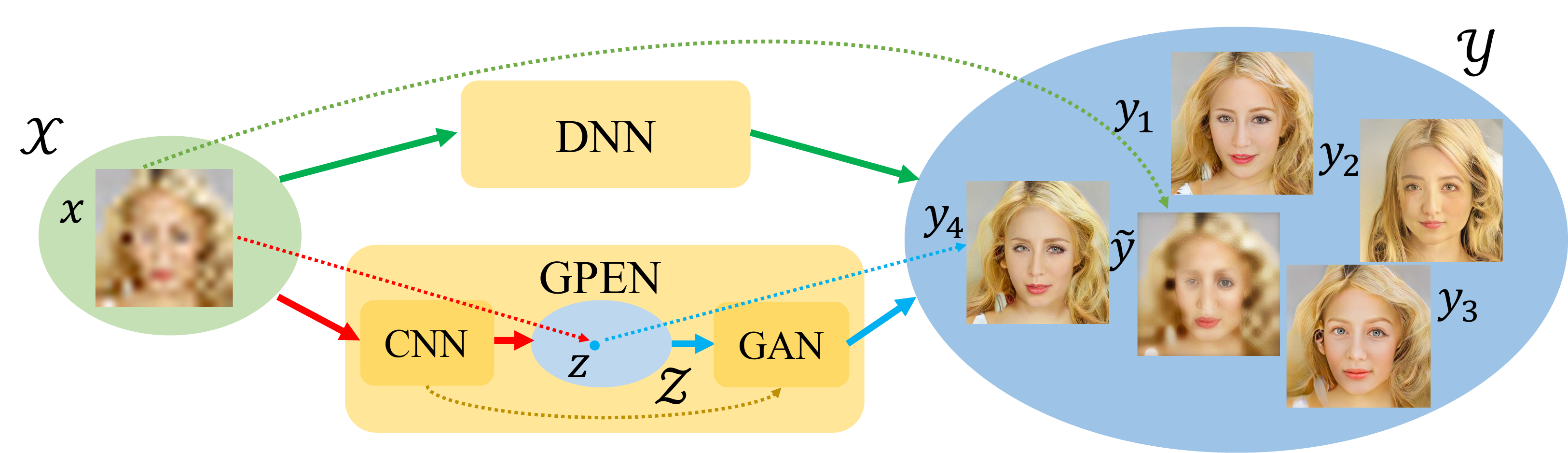}
\caption{Illustration of the motivation and framework of our GAN prior embedded network (GPEN).}
\label{fig:motivation}
\vspace*{-3mm}
\end{figure}

\begin{figure*}[t!]
\centering
\includegraphics[width=0.96\textwidth]{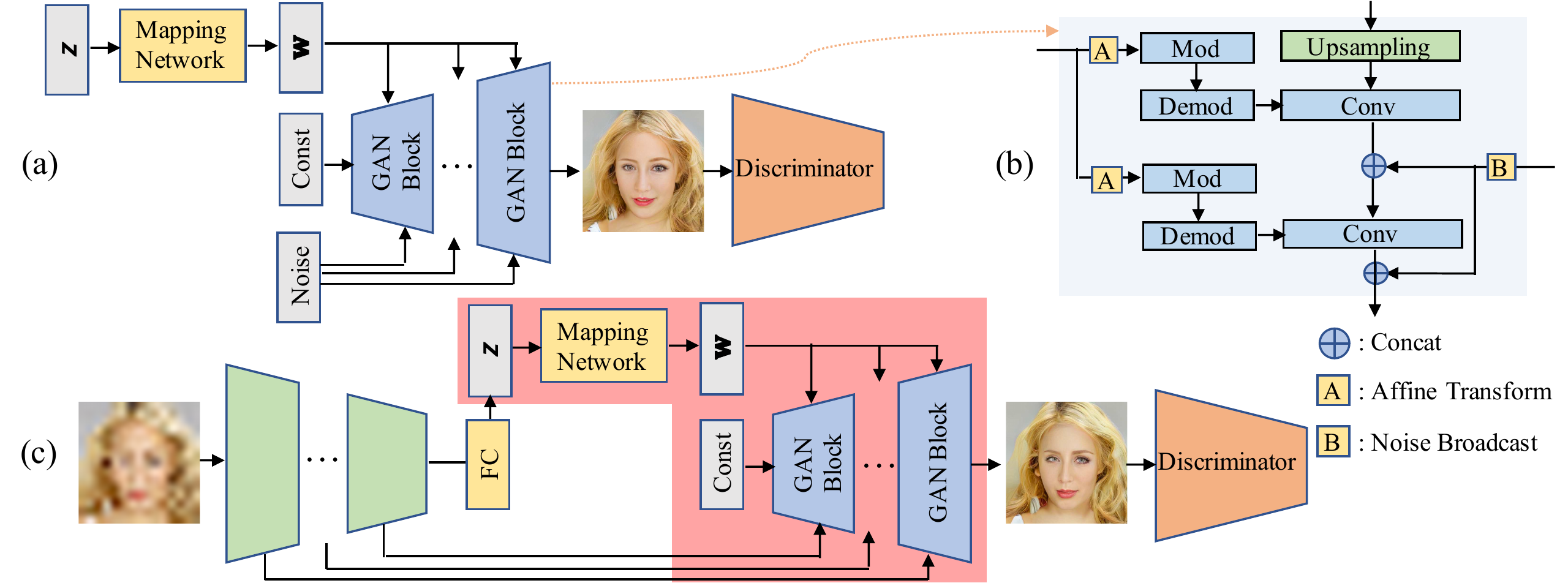}
\caption{The architecture of GPEN. (a) The GAN prior network; (b) detailed structures of a GAN block; and (c) the full network architecture of GPEN. The definition of ``Mod'' and ``Demod'' can be found in \cite{Karras2019StyleGAN2}.}
\label{fig:arch}
\vspace*{-3mm}
\end{figure*}

\section{Proposed Method}
\label{sec:proposed}

\subsection{Motivation and Framework}
\label{sec:motivation}
BFR is a typical ill-posed inverse problem. Denote by $\mathcal{X}$ the space of degraded LQ faces, and by $\mathcal{Y}$ the space of original HQ face images. Given an input LQ face image $\mathbf{x}\in\mathcal{X}$, BFR aims to find its corresponding clear face image $\mathbf{y}\in\mathcal{Y}$. Most of the DNN based methods learn a mapping function $\Phi$ to achieve this goal, i.e., $\Phi(\mathbf{x})\rightarrow\mathbf{y}$. However, this is a one-to-many inverse problem, and there are many possible face images (e.g., $\mathbf{y}_1,\mathbf{y}_2,...,\mathbf{y}_n$) in $\mathcal{Y}$ that can match to the input $\mathbf{x}$. Existing methods \cite{Bulat2018SuperFAN,Li2018GFRNet,Dogan2019Exemplar} usually train DNNs to perform mapping between $\mathbf{x}$ and $\mathbf{y}$ using some pixel-wise loss functions. As a result, as we illustrated in Figure~\ref{fig:motivation}, the final solution $\Phi(\mathbf{x})$ tends to be the mean of those HQ faces, which is over-smoothed and loses details. This coincides with the visual perception global-first theory \cite{Chen2005Global}. The cGAN methods \cite{Isola2017Pix2Pix,Wang2018Pix2PixHD} can partially dilute this issue by adversarial training to reduce the uncertainty in mapping. However, when the degradation is severe, the problem remains and cGANs can hardly generate clear face images with realistic textures and details (see Figure~\ref{fig:comp} for example).

Different from previous methods \cite{Bulat2018SuperFAN,Li2018GFRNet,Dogan2019Exemplar,Wang2018Pix2PixHD,Yang2020HiFaceGANFR}, we first train a GAN prior network, and then embed it into a DNN as decoder for HQ face image restoration. We call our method GAN prior embedded network (GPEN). As illustrated in Figure~\ref{fig:motivation}, the first part of our GPEN is a CNN encoder, which learns to map the input degraded image $\mathbf{x}$ to a desired latent code $\mathbf{z}$ in the latent space $\mathcal{Z}$ of the GAN. The GAN prior network can then reproduce the desired HQ face image via $G(\mathbf{z})\rightarrow\mathbf{y}$, where G refers to the learned generator of GAN. The generation process is basically a one-to-one mapping, largely alleviating the uncertainty of one-to-many mapping in previous methods. It should be noted that the GAN inversion methods \cite{Gu2019Prior,Menon2020PULSE,Richardson2020pSp} share a similar idea with our GPEN; however, they keep the pre-trained GANs unchanged for consistent and convenient face manipulations. While in GPEN, we carefully design and pre-train the GAN blocks and fine-tune the GAN priors for effective BFR. The architectures of GPEN and GAN blocks are shown in Figure~\ref{fig:arch} and will be explained in detail in the following sections.

\subsection{Network Architecture}
\label{sec:model}
\textbf{The GAN prior network.}
U-Net \cite{Ronneberger2015Unet} has been successfully and widely used in many image restoration tasks \cite{Wang2018Pix2PixHD,Guo2019CBDNet} and demonstrated its effectiveness in preserving image details. Therefore, our GPEN overall follows a U-shaped encoder-decoder architecture (see Figure~\ref{fig:arch}(c)). Accordingly, the GAN prior network should be designed to meet two requirements: 1) it is capable of generating HQ face images; and 2) it can be readily embedded into the U-shaped GPEN as a decoder. Inspired by the state-of-the-art GAN architectures, e.g., StyleGAN \cite{Karras2018StyleGAN,Karras2019StyleGAN2}, we use a mapping network to project latent code $\mathbf{z}$ into a less entangled space $\mathbf{w}\in{\mathcal{W}}$, as illustrated in Figure~\ref{fig:arch}(a). The intermediate code $\mathbf{w}$ is then broadcasted to each GAN block. Since the GAN prior network will be embedded into a U-shaped DNN for finetuning, we need to leave room for the skipped feature maps extracted by the encoder of the U-shaped DNN. We thus provide additional noise inputs to each GAN block. 

For the structure of GAN block, there are several options. In this work, we adopt the architecture in StyleGAN v2 (see Figure~\ref{fig:arch}(b)) due to its high capability to generate HQ images. (Alternative GAN architectures such as StyleGAN v1 \cite{Karras2018StyleGAN}, PGGAN \cite{Karras2018PGGAN} and BigGAN \cite{Brock2019BigGAN} can also be easily adopted into our GPEN.) The number of GAN blocks is equal to the number of skipped feature maps extracted in the U-shaped DNN (and the number of noise inputs), which is related to the resolution of input face image. StyleGAN requires two different noise inputs in each GAN block. To enable the GAN prior network to be readily embedded into the U-shaped GPEN, different from StyleGAN, the noise inputs are reused at the same spatial resolution for all GAN blocks. Furthermore, the noise inputs are concatenated rather than added to the convolutions in StyleGAN. We empirically found that this can bring more details in the restored face image.

\textbf{Full network architecture.}
Once the GAN prior network is trained by using some dataset (e.g., the FFHQ \cite{Karras2018StyleGAN} dataset), we embed it into the U-shaped DNN as a decoder, as shown in Figure~\ref{fig:arch}(c). The latent code $\mathbf{z}$ and the noise inputs to the GAN network are replaced by the output of the fully-connected layer (i.e., deeper features) and shallower layers of the encoder of the DNN, respectively, which will control the reconstruction of global face structure, local face details, as well as the background of face image. Since the proposed model is not fully convolutional, LQ face images are first resized to the desired resolution (e.g., $1024^2$) using simple bilinear interpolator before being input to the GPEN. After embedding, the whole GPEN will be fine-tuned so that the encoder part and decoder part can learn to adapt to each other.

\subsection{Training Strategy}
\label{sec:training}
We first pre-train the GAN prior network using a dataset of HQ face images following the training strategies of StyleGAN \cite{Karras2018StyleGAN,Karras2019StyleGAN2}. The pre-trained GAN model is embedded into the proposed GPEN, and we fine-tune the whole network using a set of synthesized LQ-HQ face image pairs (the image synthesis process will be given in Section~\ref{sec:degradation}).  

To fine-tune the GPEN model, we adopt three loss functions: the adversarial loss $L_A$, the content loss $L_C$, and the feature matching loss $L_F$. $L_A$ is inherited from the GAN prior network:
\begin{equation}
%L_A=\min_{\tilde{G}}\bigg{[}\max_DE_{(X)}\log\bigg{(}1+\exp\Big{(}{-D\big{(}\tilde{G}(\tilde{X})\big{)}}\Big{)}\bigg{)}\bigg{]},
%L_A=\min_{G}\bigg{[}\max_DE_{(X)}\log\bigg{(}1+\exp\Big{(}{-D\big{(}G(\tilde{X})\big{)}}\Big{)}\bigg{)}\bigg{]},
L_A=\min_{G}\max_DE_{(X)}\log\bigg{(}1+\exp\Big{(}{-D\big{(}G(\tilde{X})\big{)}}\Big{)}\bigg{)},
\end{equation}
where $X$ and $\tilde{X}$ denote the ground-truth HQ image and the degraded LQ one, $G$ is the generator during training, and $D$ is the discriminator. $L_C$ is defined as the $L_1$-norm distance between the final results of the generator and the corresponding ground-truth images. $L_F$ is similar to the perceptual loss \cite{Johnson2016Perceptual} but it is based on the discriminator rather than the pre-trained VGG network to fit our task. It is formulated as follows:
\begin{equation}
%L_F = \min_{\tilde{G}}\bigg{(}E_{(X)}\Big{(}\sum_{i=0}^{T}\big{\|}D^i(X)-D^i(\tilde{G}(\tilde{X}))\big{\|}_2\Big{)}\bigg{)},
%L_F = \min_{G}\bigg{(}E_{(X)}\Big{(}\sum_{i=0}^{T}\big{\|}D^i(X)-D^i(G(\tilde{X}))\big{\|}_2\Big{)}\bigg{)},
L_F = \min_{G}E_{(X)}\Big{(}\sum\nolimits_{i=0}^{T}\big{\|}D^i(X)-D^i(G(\tilde{X}))\big{\|}_2\Big{)},
\end{equation}
where $T$ is the total number of intermediate layers used for feature extraction. $D^i(X)$ is the extracted feature at the $i$-th layer of discriminator $D$. 

The final loss $L$ is as follows:
\begin{equation}
L = L_A + \alpha L_C + \beta L_F,
\end{equation}
where $\alpha$ and $\beta$ are balancing parameters. The content loss $L_C$ enforces the fine features and preserves the original color information. By introducing the feature matching loss $L_F$ on the discriminator, the adversarial loss $L_A$ can be better balanced to recover more realistic face images with vivid details. In all the following experiments, we empirically set $\alpha=1$ and $\beta=0.02$.

\section{Experiments}
\label{sec:experiments}
\subsection{Datasets and Evaluation Metric}
The FFHQ dataset \cite{Karras2018StyleGAN}, which contains $70,000$ HQ face images of resolution $1024^2$, is used to train our GPEN model. We first use it to train the GAN prior network, and then synthesize LQ images from it to fine-tune the whole GPEN. To evaluate our model, we use the CelebA-HQ dataset \cite{Karras2018PGGAN} to simulate LQ face images to quantitatively compare GPEN with other state-of-the-art methods. We also collet $1,000$ real-world LQ faces (will be made publicly available) from internet to qualitatively evaluate the performance of our model in the wild. In the quantitative evaluation, the Peak Signal-to-Noise Ratio (PSNR),  the Fr\'echet Inception Distances (FID) \cite{Heusel2017FID} and the Learned Perceptual Image Patch Similarity (LPIPS) \cite{Zhang2018LPIPS} indices are used. It is worth mentioning that all these indices can only be used as references for evaluation because they cannot truly reflect the performance of a BFR method, especially for BFR in the wild. 

\subsection{Implementation Details}
\label{sec:degradation}
We first train the GAN prior network using the FFHQ dataset with similar settings to StyleGAN \cite{Karras2018StyleGAN,Karras2019StyleGAN2}. The pre-trained GAN prior network is embedded into the GPEN to perform fine-tuning. To build LQ-HQ image pairs for fine-tuning, we synthesize degraded faces from the HQ images in FFHQ using the following degradation model:
\begin{equation}
I^d=((I\otimes\mathbf{k})\downarrow_s+\mathbf{n}_\sigma)_{{JPEG}_q},
\label{eqn:degradation}
\end{equation}
where $I$, $\mathbf{k}$, $\mathbf{n}_\sigma$, $I^d$ are respectively the input face image, the blur kernel, the Gaussian noise with standard deviation $\sigma$ and the degraded image. $\otimes$, $\downarrow_s$, ${JPEG}_q$ respectively denote the two-dimensional convolution, the standard $s$-fold downsampler and the JPEG compression operator with a quality factor $q$. 

The above degradation model has been used in previous methods \cite{Li2018GFRNet,Li2020Restore}. In our implementation, for each image the blur kernel $\mathbf{k}$ is randomly selected from a set of blurring models, including Gaussian blur and motion blur with varying kernel sizes. The additive Gaussian noise $\mathbf{n}_\sigma$ is sampled channel-wise from a normal distribution, and $\sigma$ is chosen from $[0, 25]$. The value of $s$ is randomly and uniformly sampled from $[10, 200]$ (i.e., up to $200$ times downscaling) and $q$ is randomly and uniformly sampled from $[5, 50]$ (i.e., up to $95\%$ JPEG compression) per image. By using those severely degraded images to fine-tune the model, the encoder part of our GPEN can learn to generate suitable latent code and noise inputs to the GAN prior decoder network, which is updated simultaneously to tackle severely degraded faces in real-world scenarios. 

During model updating, we adopt the Adam optimizer with a batch size of $1$. The learning rate ($LR$) varies for different parts of GPEN, including the encoder, the decoder and the discriminator. In our implementation, we let $LR_{\text{encoder}}=0.002$, and set $LR_{\text{encoder}}:LR_{\text{decoder}}:LR_{\text{discriminator}}=100:10:1$. It should be noted that the discriminator part will be removed in the testing stage.

\subsection{Ablation Study}
To better understand the roles of different components of GPEN and the training strategy, in this section we conduct an ablation study by introducing some variants of GPEN and comparing their BFR performance. The first variant is denoted by GPEN-w/o-ft, i.e., the embedded GAN prior network is kept unchanged in the fine-tuning process. The second variant is denoted by GPEN-w/o-noise, which refers to the GPEN model without noise inputs. The third variant is denoted by GPEN-noise-add, i.e., that the noise inputs are added rather than concatenated to the convolutions.
%To better understand the roles of different components of GPEN and the training strategy, in this section we conduct an ablation study by introducing some variants of GPEN and comparing their BFR performance. The first variant is denoted by GPEN-w/o-ft, i.e., the embedded GAN prior network is kept unchanged in the fine-tuning process. The second variant is denoted by GPEN-w/o-noise, which refers to the GPEN model without noise inputs. The third variant is denoted by GPEN-noise, i.e., that the noise inputs to the GAN prior network are set as random noise but not the skipped features of encoder network. The fourth variant is denoted by GPEN-noise-add, i.e., that the noise inputs are added rather than concatenated to the convolutions.

\begin{figure*}[ht!]
\centering
    \begin{subfigure}[t!]{.16\textwidth}
        \includegraphics[width=\textwidth]{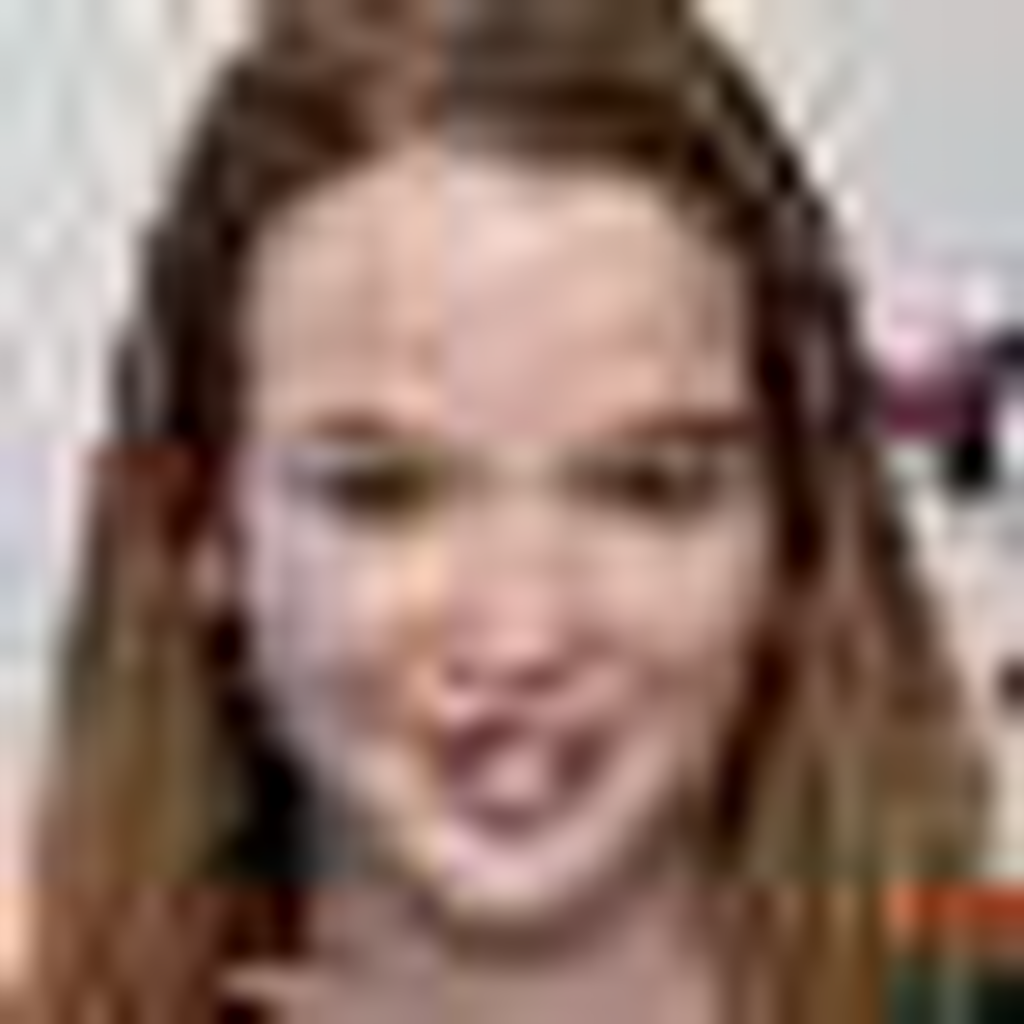}
        \vspace*{-5mm}
        \caption{}
    \end{subfigure} 
    \begin{subfigure}[t!]{.16\textwidth}
        \includegraphics[width=\textwidth]{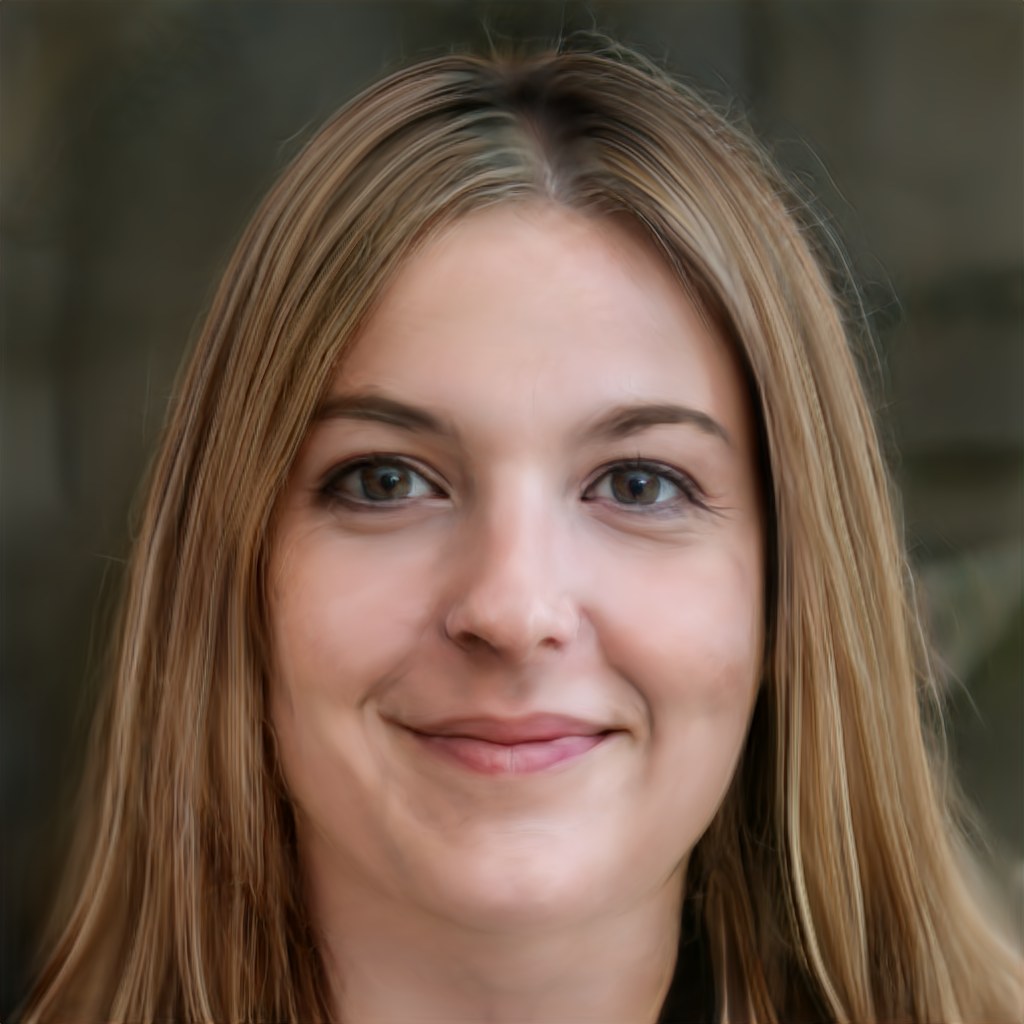}
        \vspace*{-5mm}
        \caption{}
    \end{subfigure} 
    \begin{subfigure}[t!]{.16\textwidth}
        \includegraphics[width=\textwidth]{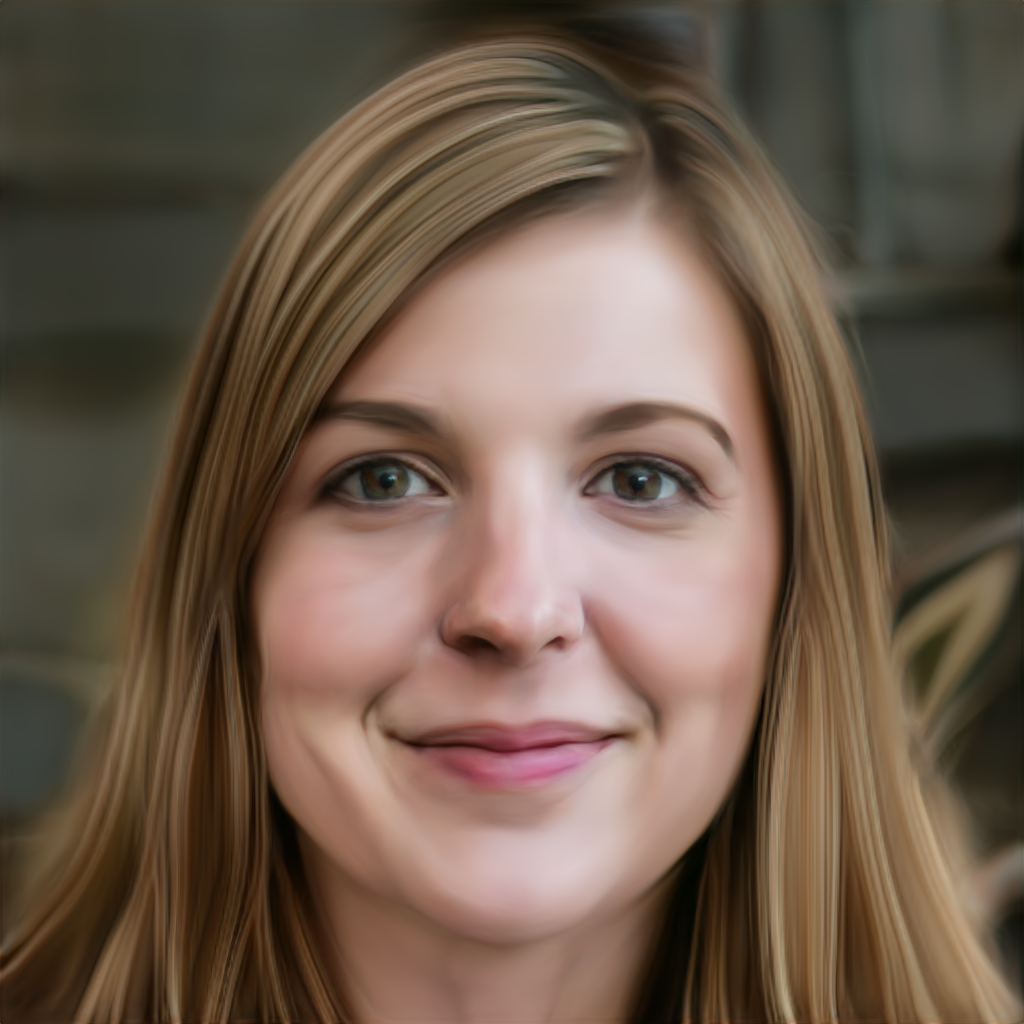}
        \vspace*{-5mm}
        \caption{}
    \end{subfigure} 
    %\begin{subfigure}[t!]{.13\textwidth}
    %    \includegraphics[width=\textwidth]{imgs/face/ablation/00030_noise.png}
    %    \caption{}
    %\end{subfigure} 
    \begin{subfigure}[t!]{.16\textwidth}
        \includegraphics[width=\textwidth]{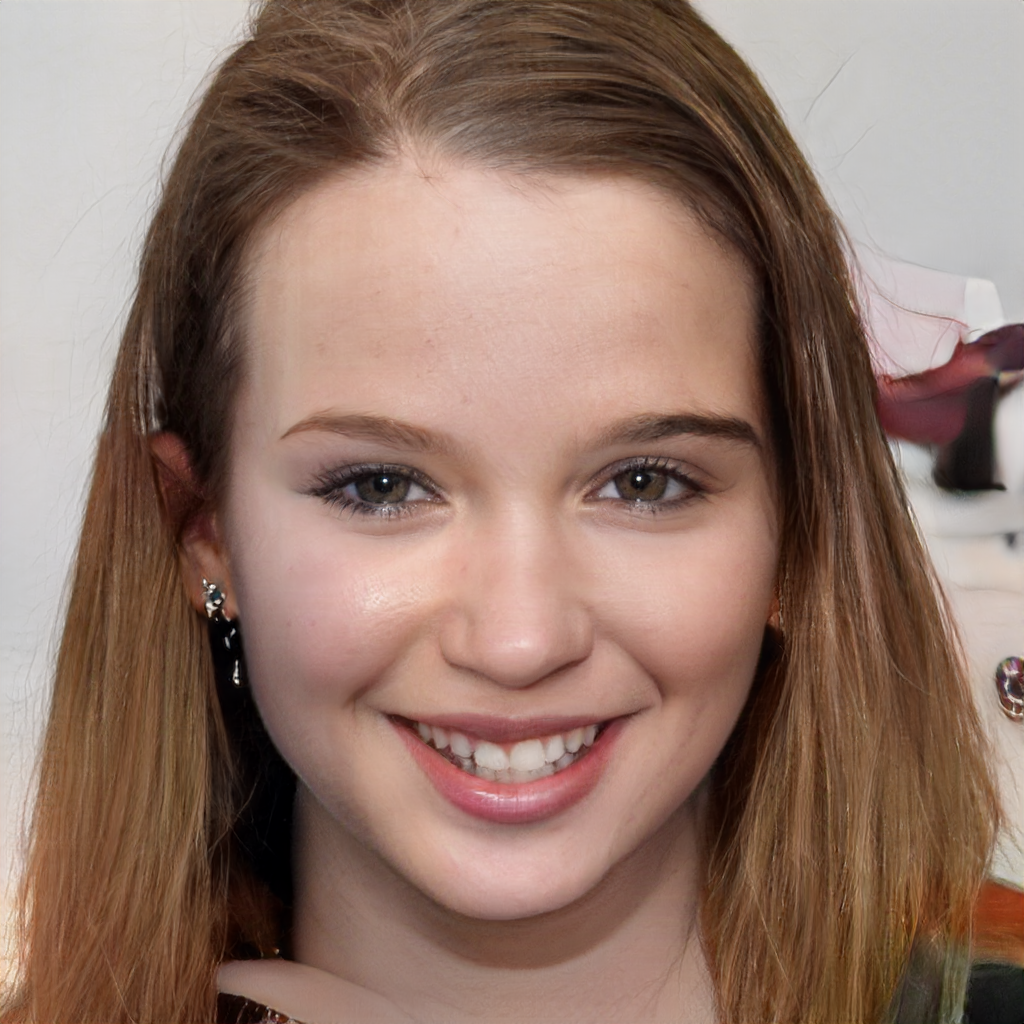}
        \vspace*{-5mm}
        \caption{}
    \end{subfigure} 
    \begin{subfigure}[t!]{.16\textwidth}
        \includegraphics[width=\textwidth]{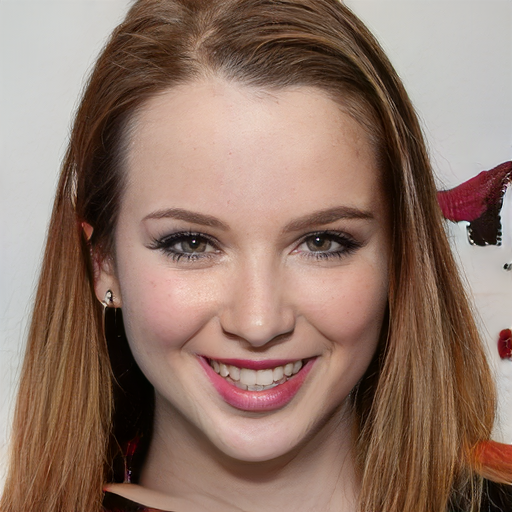}
        \vspace*{-5mm}
        \caption{}
    \end{subfigure} 
    \begin{subfigure}[t!]{.16\textwidth}
        \includegraphics[width=\textwidth]{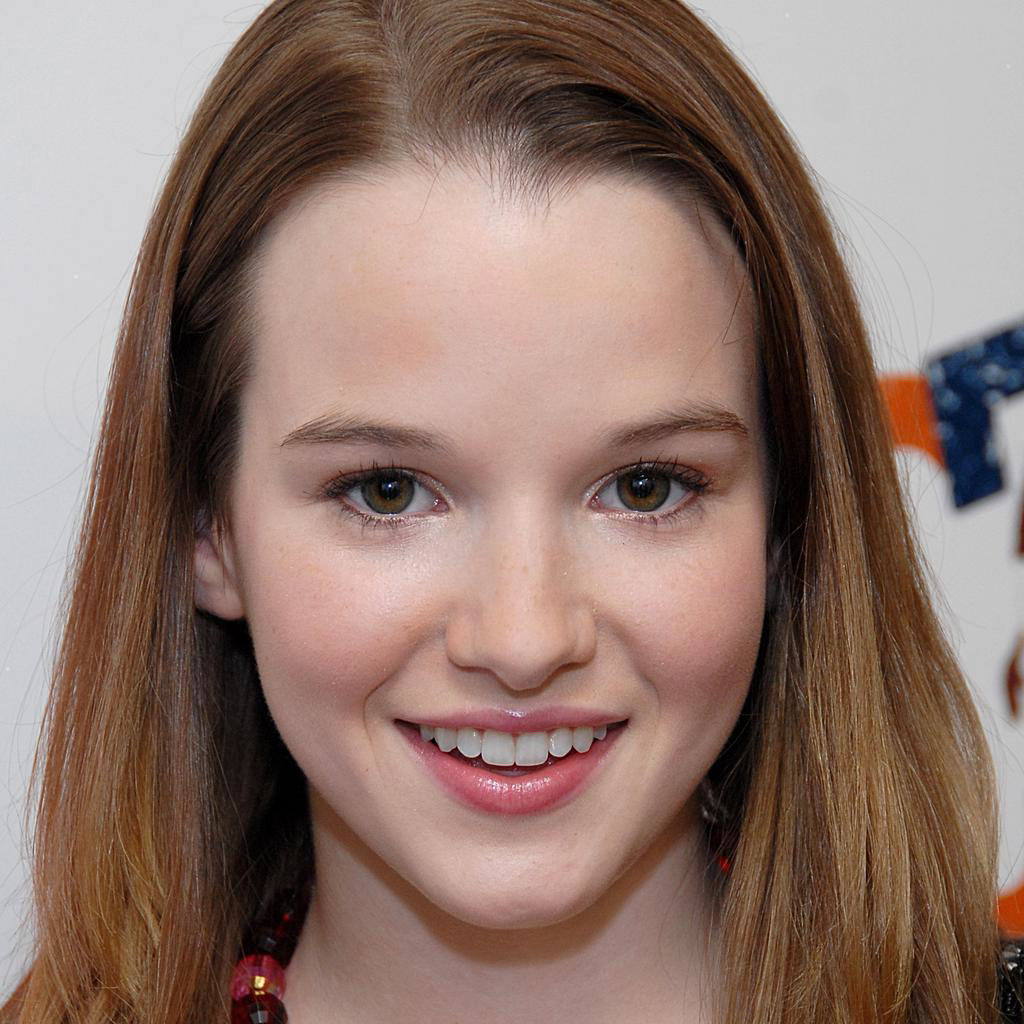}
        \vspace*{-5mm}
        \caption{}
    \end{subfigure} 
\vspace*{-3mm}
\caption{Comparisons of our variants BFR. (a) LQ input; (b) GPEN-w/o-ft; (c) GPEN-w/o-noise; (d) GPEN-noise-add; (e) GPEN; (f) Ground truth.} 
\label{fig:ablation}
\vspace*{-3mm}
\end{figure*}

\begin{figure*}[ht!]
\centering
    %\begin{subfigure}[t!]{.11\textwidth}
    %    \includegraphics[width=\textwidth]{imgs/face/syn/00304_input.png}
    %\end{subfigure}%
    %\begin{subfigure}[t!]{.11\textwidth}
    %    \includegraphics[width=\textwidth]{imgs/face/syn/00304_SuperFAN.png}
    %\end{subfigure}%
    %\begin{subfigure}[t!]{.11\textwidth}
    %    \includegraphics[width=\textwidth]{imgs/face/syn/00304_GFRNet.png}
    %\end{subfigure}%
    %\begin{subfigure}[t!]{.11\textwidth}
    %    \includegraphics[width=\textwidth]{imgs/face/syn/00304_GWAInet.png}
    %\end{subfigure}%
    %\begin{subfigure}[t!]{.11\textwidth}
    %    \includegraphics[width=\textwidth]{imgs/face/syn/00304_pix2pixHD.jpg}
    %\end{subfigure}%
    %\begin{subfigure}[t!]{.11\textwidth}
    %    \includegraphics[width=\textwidth]{imgs/face/syn/00304_DFDNet.png}
    %\end{subfigure}%
    %\begin{subfigure}[t!]{.11\textwidth}
    %    \includegraphics[width=\textwidth]{imgs/face/syn/00304_HiFaceGAN.jpg}
    %\end{subfigure}%
    %\begin{subfigure}[t!]{.11\textwidth}
    %    \includegraphics[width=\textwidth]{imgs/face/syn/00304_ours.png}
    %\end{subfigure}%
    %\begin{subfigure}[t!]{.11\textwidth}
    %    \includegraphics[width=\textwidth]{imgs/face/syn/00304_gt.png}
    %\end{subfigure}%
    %\\
    \begin{subfigure}[t!]{.11\textwidth}
        \includegraphics[width=\textwidth]{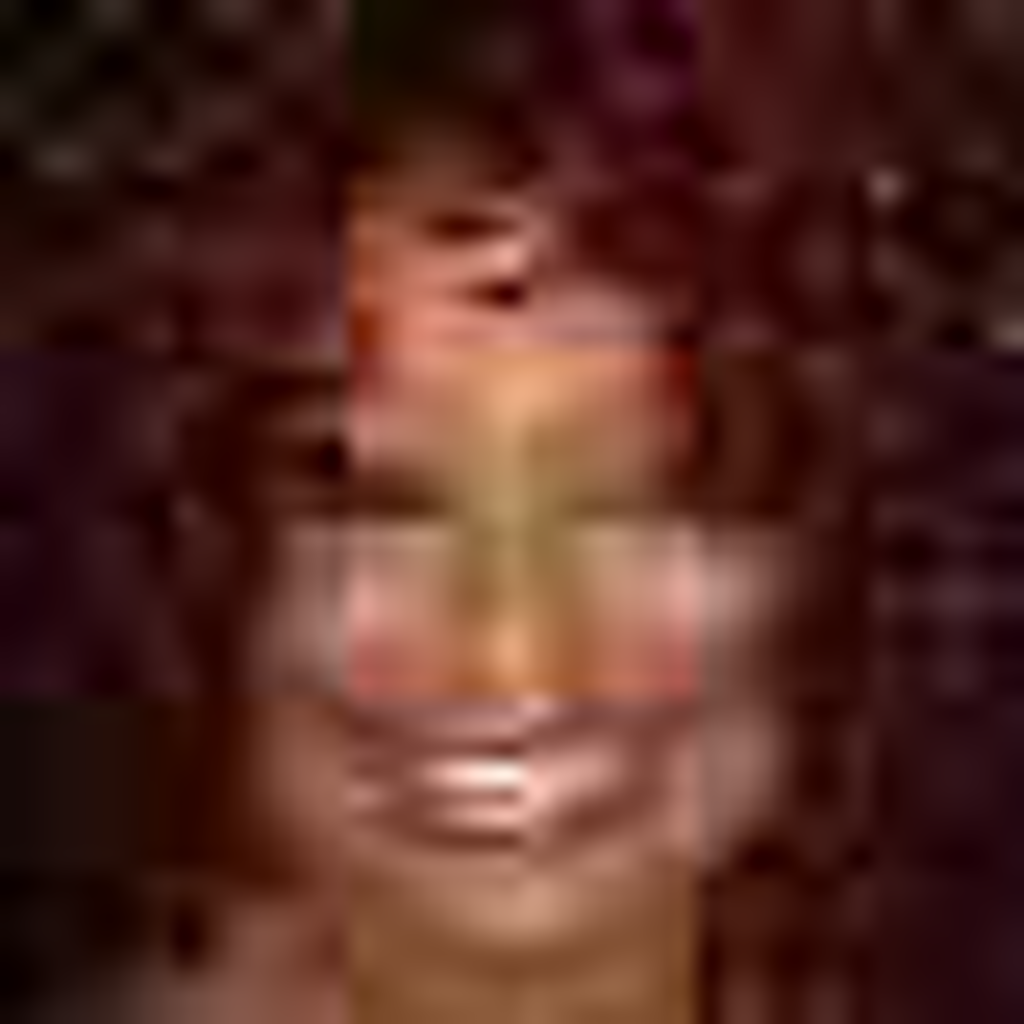}
    \end{subfigure}%
    \begin{subfigure}[t!]{.11\textwidth}
        \includegraphics[width=\textwidth]{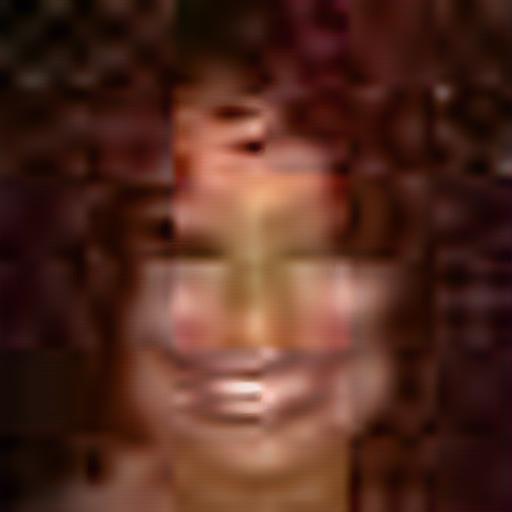}
    \end{subfigure}%
    \begin{subfigure}[t!]{.11\textwidth}
        \includegraphics[width=\textwidth]{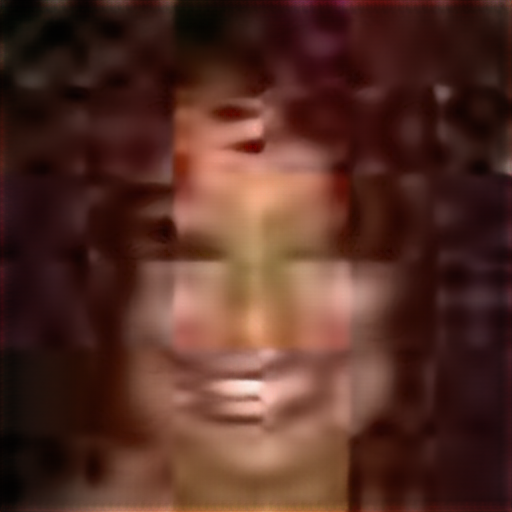}
    \end{subfigure}%
    \begin{subfigure}[t!]{.11\textwidth}
        \includegraphics[width=\textwidth]{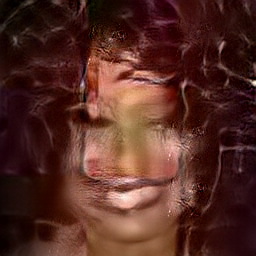}
    \end{subfigure}%
    \begin{subfigure}[t!]{.11\textwidth}
        \includegraphics[width=\textwidth]{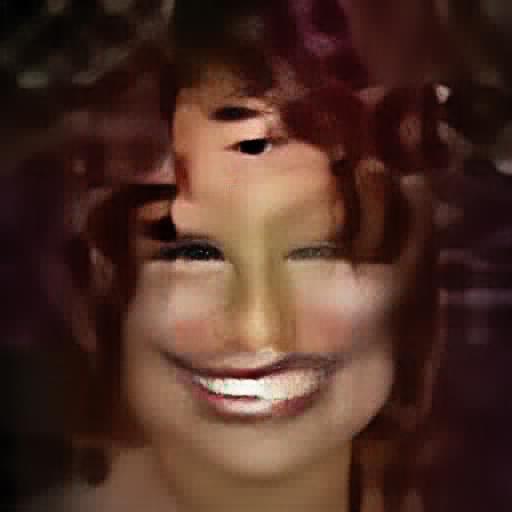}
    \end{subfigure}%
    \begin{subfigure}[t!]{.11\textwidth}
        \includegraphics[width=\textwidth]{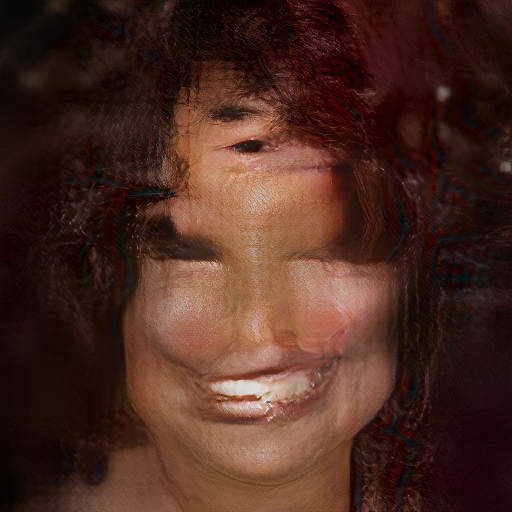}
    \end{subfigure}%
    \begin{subfigure}[t!]{.11\textwidth}
        \includegraphics[width=\textwidth]{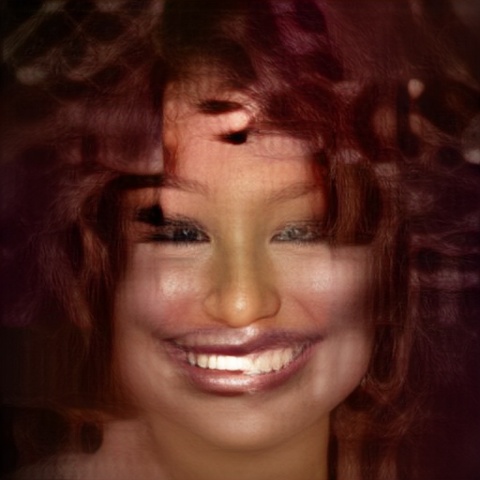}
    \end{subfigure}%
    \begin{subfigure}[t!]{.11\textwidth}
        \includegraphics[width=\textwidth]{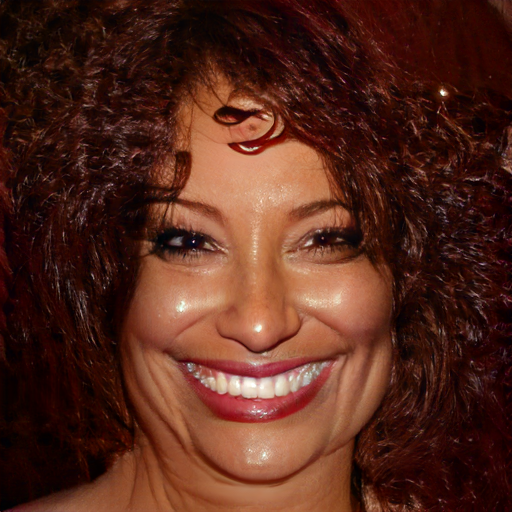}
    \end{subfigure}%
    \begin{subfigure}[t!]{.11\textwidth}
        \includegraphics[width=\textwidth]{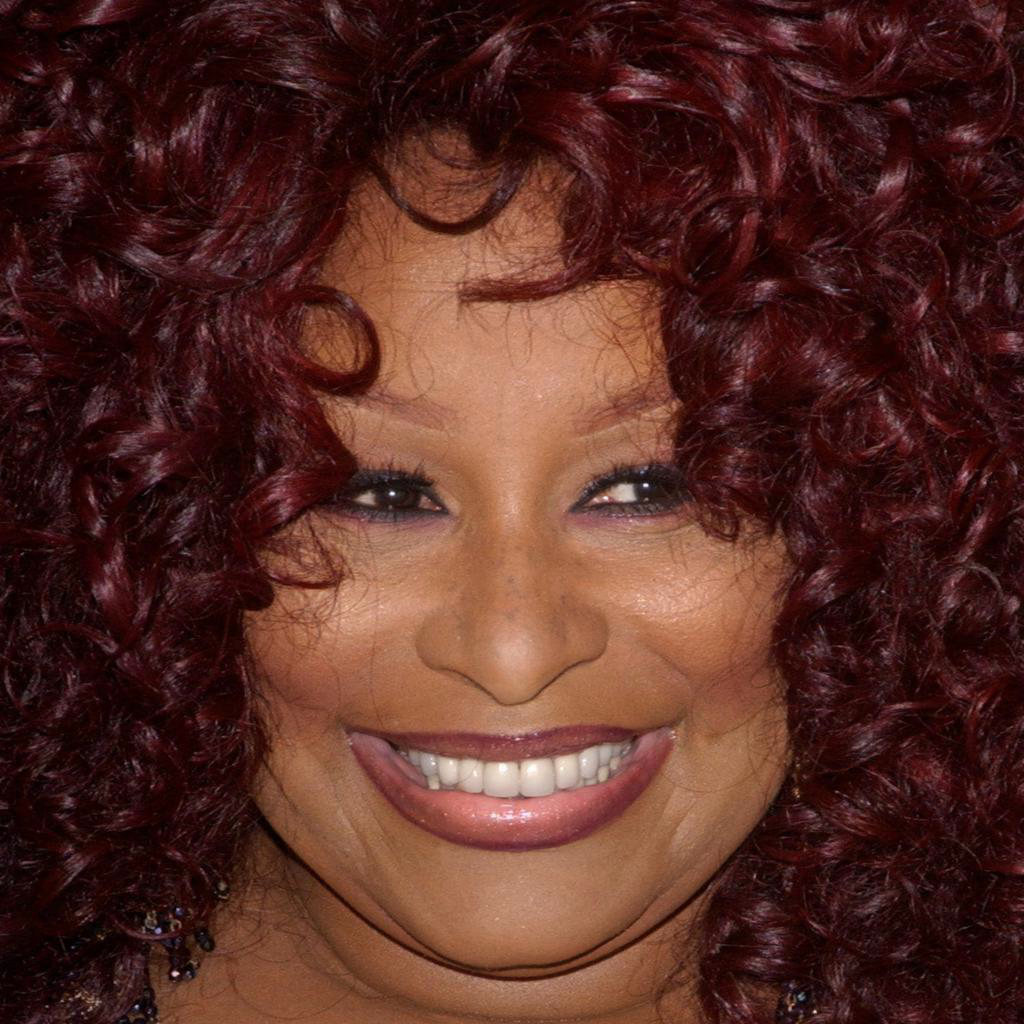}
    \end{subfigure}%
    \\
    \begin{subfigure}[t!]{.11\textwidth}
        \includegraphics[width=\textwidth]{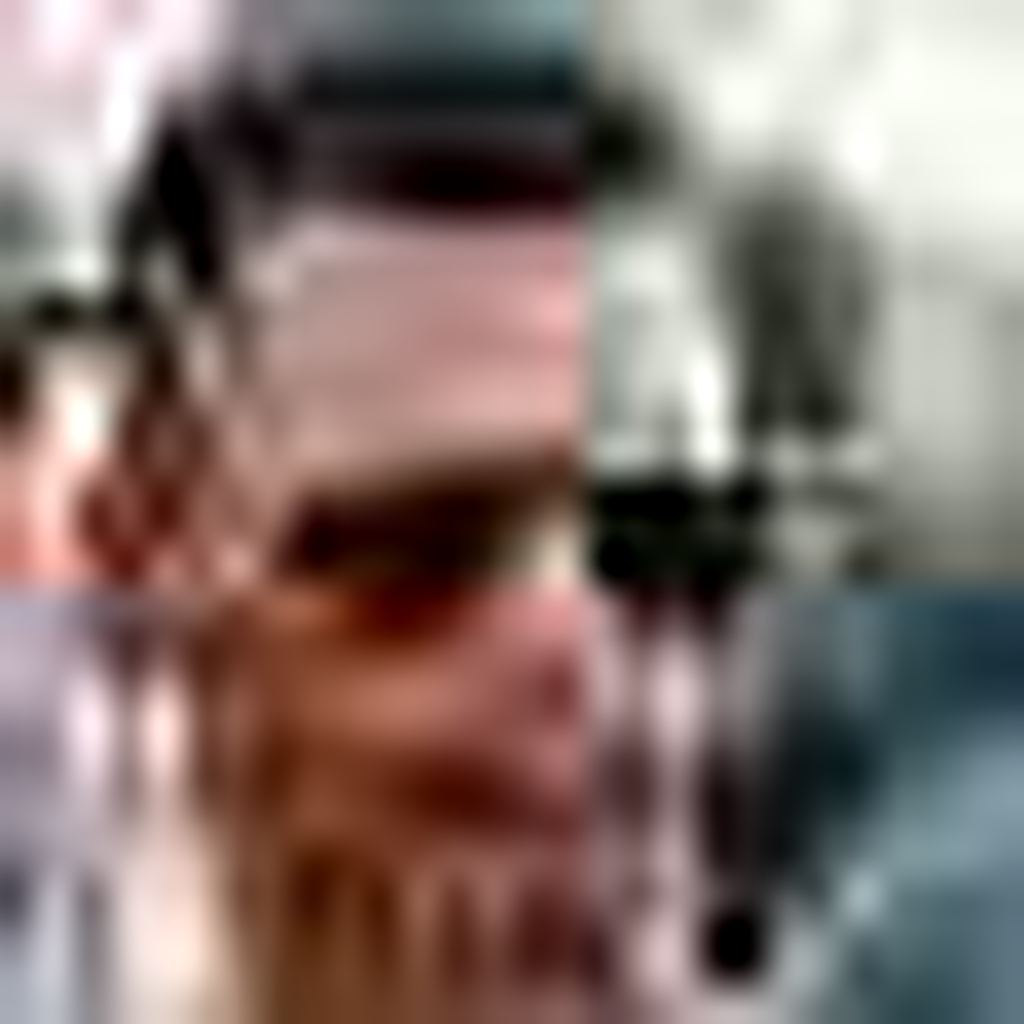}
        \vspace*{-5mm}
        \caption{}
    \end{subfigure}%
    \begin{subfigure}[t!]{.11\textwidth}
        \includegraphics[width=\textwidth]{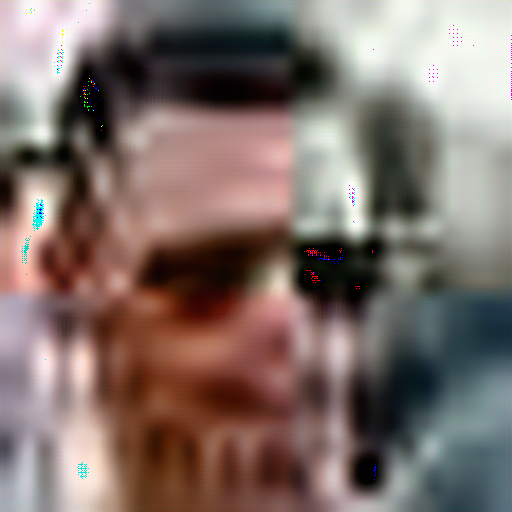}
        \vspace*{-5mm}
        \caption{}
    \end{subfigure}%
    \begin{subfigure}[t!]{.11\textwidth}
        \includegraphics[width=\textwidth]{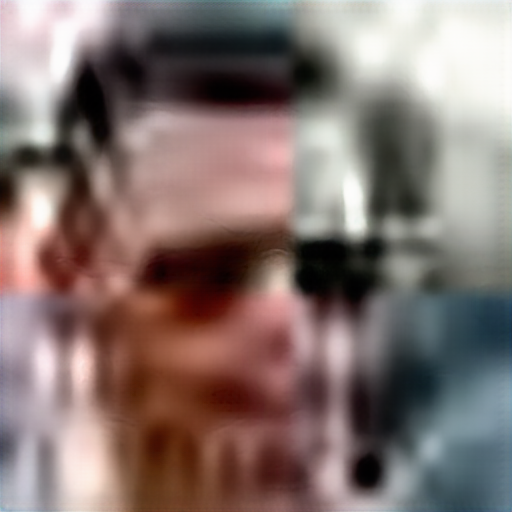}
        \vspace*{-5mm}
        \caption{}
    \end{subfigure}%
    \begin{subfigure}[t!]{.11\textwidth}
        \includegraphics[width=\textwidth]{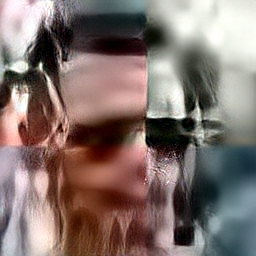}
        \vspace*{-5mm}
        \caption{}
    \end{subfigure}%
    \begin{subfigure}[t!]{.11\textwidth}
        \includegraphics[width=\textwidth]{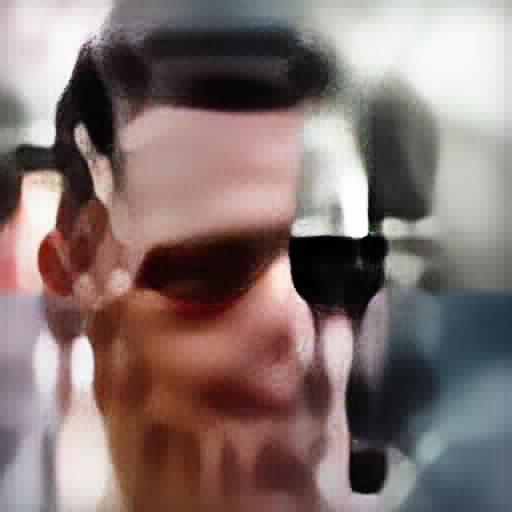}
        \vspace*{-5mm}
        \caption{}
    \end{subfigure}%
    \begin{subfigure}[t!]{.11\textwidth}
        \includegraphics[width=\textwidth]{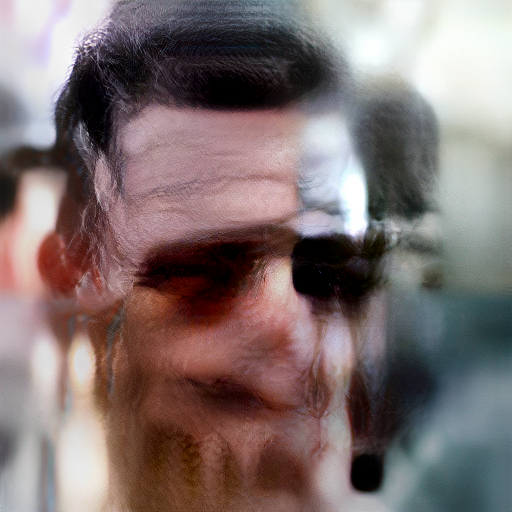}
        \vspace*{-5mm}
        \caption{}
    \end{subfigure}%
    \begin{subfigure}[t!]{.11\textwidth}
        \includegraphics[width=\textwidth]{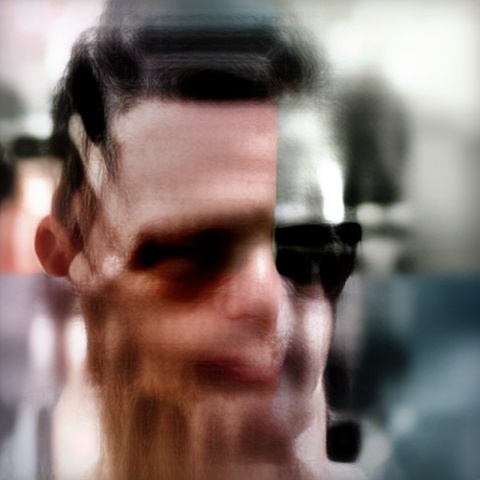}
        \vspace*{-5mm}
        \caption{}
    \end{subfigure}%
    \begin{subfigure}[t!]{.11\textwidth}
        \includegraphics[width=\textwidth]{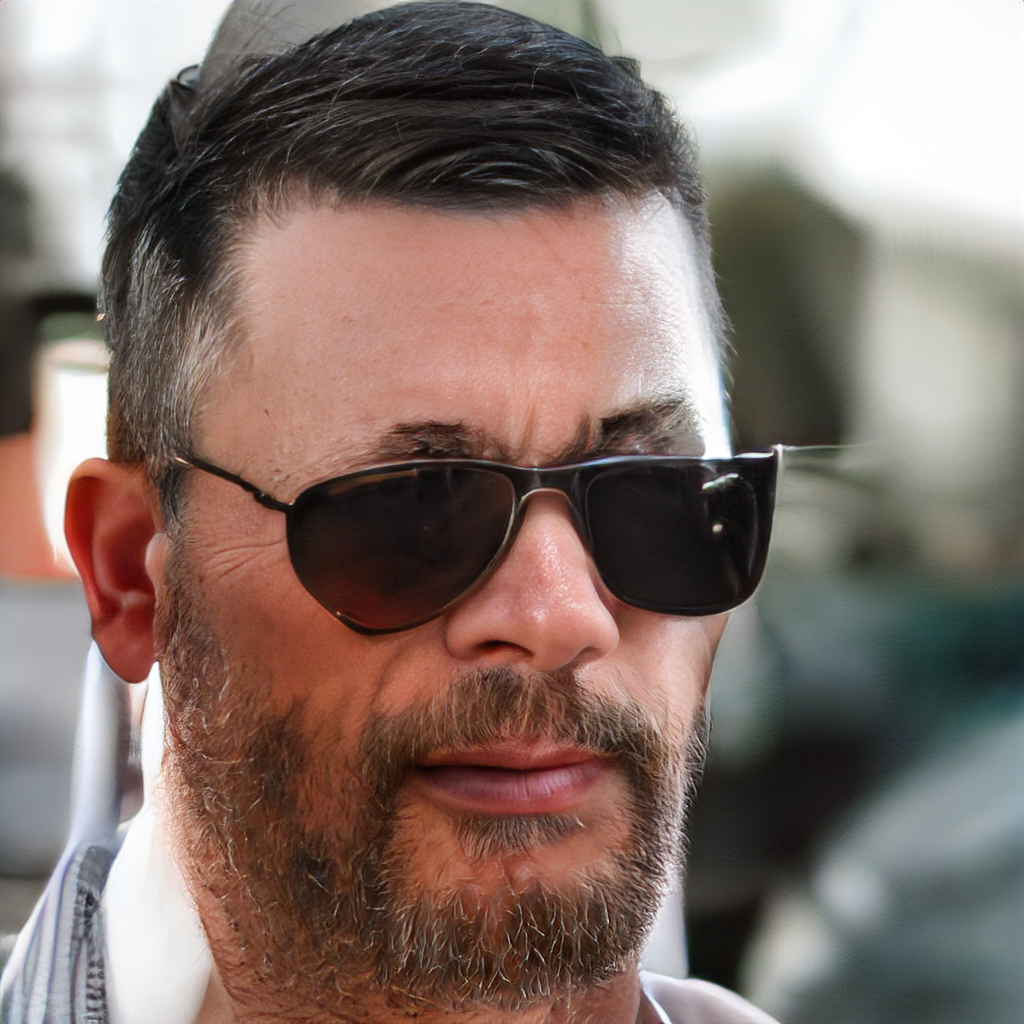}
        \vspace*{-5mm}
        \caption{}
    \end{subfigure}%
    \begin{subfigure}[t!]{.11\textwidth}
        \includegraphics[width=\textwidth]{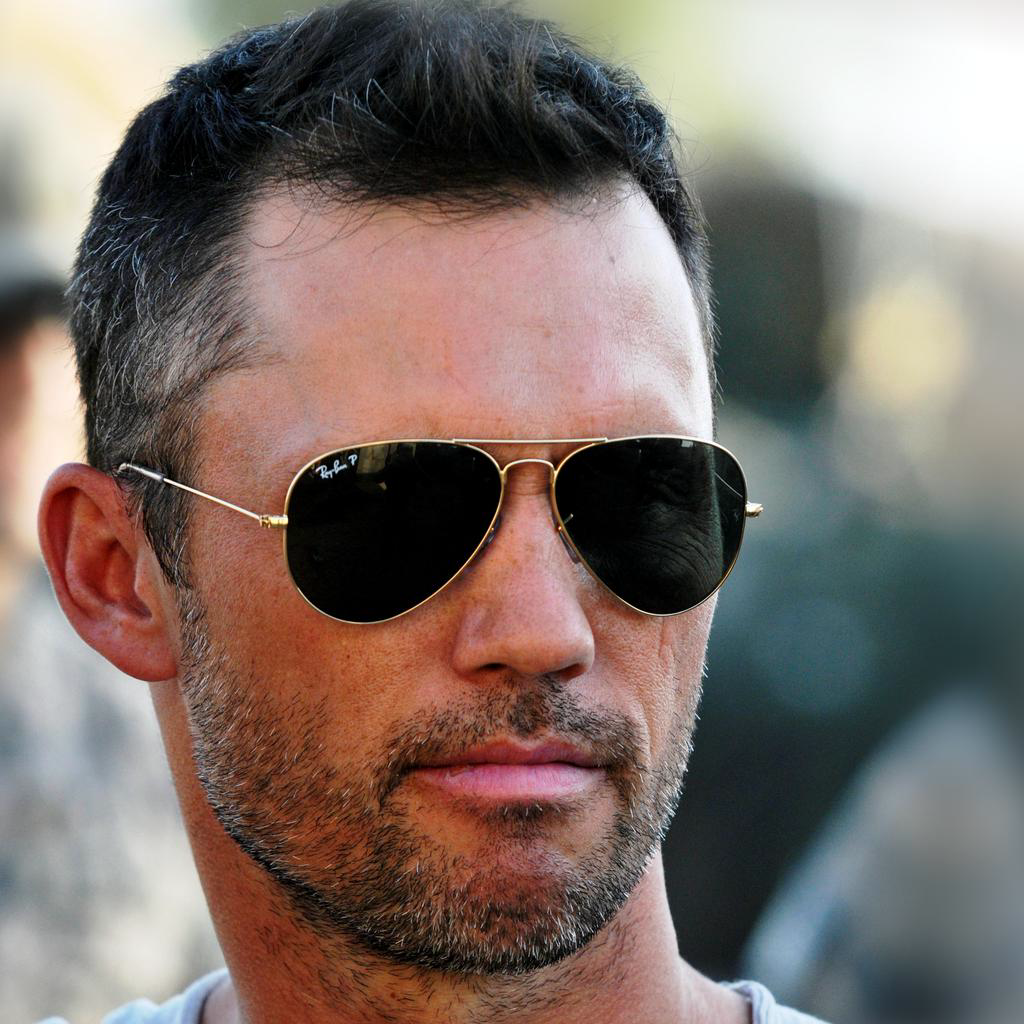}
        \vspace*{-5mm}
        \caption{}
    \end{subfigure}%
\vspace*{-3mm}
\caption{Blind face restoration results on synthsized degraded faces. (a) Degraded faces; (b) Super-FAN \protect{\cite{Bulat2018SuperFAN}}; (c) GFRNet \protect{\cite{Li2018GFRNet}}; (d) GWAInet \protect{\cite{Dogan2019Exemplar}}; (e) Pix2PixHD \protect{\cite{Wang2018Pix2PixHD}}; (f) DFDNet \protect{\cite{Li2020Restore}}; (g) HiFaceGAN \protect{\cite{Yang2020HiFaceGANFR}}; (h) GPEN; (i) Ground truth. }
%Our synthsized faces are severely degraded. Restoring these faces is beyond the capabilities of many state-of-the-art methods. Our model can perform robustly and reconstruct high-quality results with faithful details.}
%Please pay special attention to the hair, the eye, the eyebrow, the tooth, the mustache and the earring. Our method can achieve much more visually photo-realistic results. Best viewed by zooming to 400\% plus in the screen.}
\label{fig:comp}
\vspace*{-3mm}
\end{figure*}

\begin{table}[t!]
\centering
\caption{Comparison (PSNR, FID and LPIPS) of different variants of GPEN.}
\vspace*{-3mm}
   \begin{tabular}{l|c c c}
      \textbf{Method} & \textbf{PSNR}$\uparrow$ & \textbf{FID}$\downarrow$ & \textbf{LPIPS}$\downarrow$ \\
      \hline
      %Ours-v1 & 20.68 & 40.08 & 0.380 \\
      GPEN-w/o-ft & 12.55 & 92.71 & 0.653 \\
      GPEN-w/o-noise & 13.30 & 95.62 & 0.709 \\
      %GPEN-noise & - & - & - \\
      GPEN-noise-add & 20.71 & 34.26 & 0.359 \\
      GPEN & $\mathbf{20.80}$ & $\mathbf{31.72}$ & $\mathbf{0.346}$ \\
   \end{tabular}
\label{tab:ablation}
\vspace*{-5mm}
\end{table}

We perform BFR on the CelebA-HQ dataset to evaluate GPEN and its three variants. The LQ images are synthesized by using the degradation model in Eq.~(\ref{eqn:degradation}) and the same set of parameters used in Section~\ref{sec:degradation}. Table~\ref{tab:ablation} lists the PSNR, FID and LPIPS results. One can see that GPEN achieves better quantitative measures than its variants. Figure~\ref{fig:ablation} shows the BSR results of the networks on an image. We can see that GPEN-w/o-ft can generate clean HQ face image; however, the appearance of the face is rather different from the ground-truth, and the background of the image is totally different. This is because without fine-tuning the GAN prior, it is difficult to generate the desired latent code into the latent space $\mathcal{Z}$, which coincides with the findings in many GAN inversion works \cite{Abdal2019Img2StyleGAN, Richardson2020pSp}. By discarding the noise input, the result of GPEN-w/o-noise is blurrier than GPEN-w/o-ft, and there are some artifact generated in the boundary of the image. This implies that the noise input plays an import role in synthesizing localize details. GPEN-noise-add achieves comparable result to GPEN but with slightly less facial details, while it generates some false details in the background of the image. Overall, GPEN shows superior performance to its variants, demonstrating the effectiveness of concatenated U-shaped architecture and our training strategy for the BFR tasks.

\begin{table}[t!]
    \centering
    \caption{Comparison (PSNR, FID and LPIPS) of different BFR methods. \protect\footnotemark}
    \vspace*{-3mm}
    \label{tab:restore}
    %\begin{subtable}{.4\linewidth}
    %    \centering
    %    \caption{Comparison on PSNR, FID and LPIPS values on blind face restoration}
    %    \label{subtab:restore}
    %    \begin{tabular}{l|c c c}
    %    \textbf{Method} & \textbf{PSNR} & \textbf{FID} & \textbf{LPIPS} \\
    %    \hline
    %    Pix2PixHD\protect{\cite{Wang2018Pix2PixHD}} & ** & ** & ** \\
    %    DFDNet\protect{\cite{Li2020Restore}} & ** & ** & ** \\
    %    HiFaceGAN\protect{\cite{Yang2020HiFaceGANFR}} & $\mathbf{**}$ & ** & ** \\
    %    Ours & ** & $\mathbf{**}$ & ** \\
    %    \end{tabular}
    %\end{subtable}%
    %\hspace*{1 cm}
    %\begin{subtable}{.6\linewidth}
    %    \centering
    %    \caption{Comparison on PSNR, FID and LPIPS values on face super resolution}
    %    \label{subtab:sr}
    %    \begin{tabular}{l|c c c}
    %    \textbf{Method} & \textbf{PSNR} & \textbf{FID} & \textbf{LPIPS} \\
    %    \hline
    %    Bilinear & ** & ** & ** \\
    %    mGANprior\protect{\cite{Gu2019Prior}} & ** & ** & ** \\
    %    PULSE\protect{\cite{Menon2020PULSE}} & $\mathbf{**}$ & ** & ** \\
    %    Ours & ** & $\mathbf{**}$ & ** \\
    %    \end{tabular}
    %\end{subtable} 
    \begin{tabular}{l|c c c}
      \textbf{Method} & \textbf{PSNR}$\uparrow$ & \textbf{FID}$\downarrow$ & \textbf{LPIPS}$\downarrow$ \\
      \hline
      Pix2PixHD \protect{\cite{Wang2018Pix2PixHD}} & 20.45 & 76.89 & 0.494 \\
      Super-FAN \protect{\cite{Bulat2018SuperFAN}} & 21.56 & 136.83 & 0.616 \\
      %WaveletSRNet\protect{\cite{Huang2017Wavelet}} & 20.98 & 127.31 & 0.550 \\
      GFRNet \protect{\cite{Li2018GFRNet}} & $\mathbf{21.70}$ & 134.92 & 0.597 \\
      GWAInet \protect{\cite{Dogan2019Exemplar}} & 19.84 & 135.84 & 0.569 \\
      %DFDNet \protect{\cite{Li2020Restore}} & - & - & - \\
      HiFaceGAN \protect{\cite{Yang2020HiFaceGANFR}} & 21.33 & 56.67 & 0.392 \\
      GPEN & 20.80 & $\mathbf{31.72}$ & $\mathbf{0.346}$ \\
   \end{tabular}
\vspace*{-5mm}
\end{table}
\footnotetext{Note that the results of DFDNet \protect\cite{Li2020Restore} are not reported because it fails to recover many face images in this experiment.}

\begin{table*}
\centering
\caption{Comparison (PSNR, FID and LPIPS) of various FSR methods. Since mGANprior \protect{\cite{Gu2019Prior}} and PULSE \protect{\cite{Menon2020PULSE}} are very time-consuming, we only used the first $1,000$ images of CelebA-HQ dataset to compute their measures. ``-'' means that the result is not available.}
\vspace*{-3mm}
    \resizebox{\textwidth}{!}{\begin{tabular}{l|c c c c c c|c c c c c c|c c c c c c}
      \multirow{2}{*}{\textbf{Method}} & 
      \multicolumn{6}{c}{\textbf{PSNR}$\uparrow$} & 
      \multicolumn{6}{c}{\textbf{FID}$\downarrow$} & 
      \multicolumn{6}{c}{\textbf{LPIPS}$\downarrow$} \\
      & $8\times$ & $16\times$ & $32\times$ & $64\times$ & $128\times$ & $256\times$ & $8\times$ & $16\times$ & $32\times$ & $64\times$ & $128\times$ & $256\times$ & $8\times$ & $16\times$ & $32\times$ & $64\times$ & $128\times$ & $256\times$ \\
      \hline
      Bilinear & $\mathbf{28.73}$ & $\mathbf{26.13}$ & $\mathbf{22.81}$ & $\mathbf{20.49}$ & $\mathbf{17.75}$ & $\mathbf{15.17}$ & 89.29 & 183.50 & 206.03 & 342.63 & 528.17 & 495.03 & 0.471 & 0.567 & 0.659 & 0.713 & 0.765 & 0.812 \\
      Super-FAN \protect{\cite{Bulat2018SuperFAN}} & - & 20.95 & - & - & - & - & - & 92.65 & - & - & - & - & - & 0.453 & - & - & - & - \\
      %WaveletSRNet\protect{\cite{Huang2017Wavelet}} & 29.02 & 26.75 & 22.53 & - & - & - & - & 28.16 & 61.17 & 88.40 & - & - & - & - & 0.228 & 0.376 & 0.429 & - & - & - & - \\
      GFRNet \protect{\cite{Li2018GFRNet}} & 28.08 & 24.73 & 21.39 & - & - & - & 47.38 & 70.49 & 132.88 & - & - & - & 0.324 & 0.423 & 0.578 & - & - & - \\
      GWAInet \protect{\cite{Dogan2019Exemplar}} & 25.79 & - & - & - & - & - & 56.81 & - & - & - & - & - & 0.339 & - & - & - & - & - \\
      DFDNet \protect{\cite{Li2020Restore}} & 25.37 & 23.11 & - & - & - & - & 29.97 & 35.46 & - & - & - & - & 0.212 & 0.274 & - & - & - & - \\
      %HiFaceGAN\protect{\cite{Yang2020HiFaceGANFR}} & 26.98 & 26.36 & 24.66 & 22.42 & 19.83 & 17.00 & 14.19 & 32.47 & $\mathbf{29.95}$ & 36.26 & 47.17 & 88.28 & 265.57 & 359.23 & 0.184 & 0.211 & 0.266 & 0.349 & 0.460 & 0.619 & 0.707 \\
      HiFaceGAN \protect{\cite{Yang2020HiFaceGANFR}} & 26.36 & 24.66 & 22.42 & 19.83 & - & - & $\mathbf{29.95}$ & 36.26 & 47.17 & 88.28 & - & - & 0.211 & 0.266 & 0.349 & 0.460 & - & - \\
      mGANprior \protect{\cite{Gu2019Prior}} & 21.44 & 21.29 & 20.53 & 18.09 & 15.45 & 13.39 & 104.20 & 100.84 & 95.82 & 108.05 & 113.73 & 113.28 & 0.521 & 0.518 & 0.472 & 0.519 & 0.558 & 0.582 \\
      PULSE \protect{\cite{Menon2020PULSE}} & 24.32 & 22.54 & 19.98 & 16.09 & 13.39 & 11.49 & 65.89 & 65.33 & 81.23 & 87.45 & 102.48 & 101.35 & 0.421 & 0.425 & 0.405 & 0.492 & 0.544 & 0.579 \\
      pSp \protect{\cite{Richardson2020pSp}} & 18.99 & 18.73 & 18.62 & 18.02 & 16.18 & 14.57 & 40.97 & 43.37 & 75.92 & 74.46 & 88.44 & 123.85 & 0.415 & 0.424 & 0.441 & 0.458 & 0.504 & 0.581 \\
      GPEN & 24.66 & 23.27 & 21.23 & 19.02 & 15.74 & 13.66 & 30.49 & $\mathbf{31.37}$ & $\mathbf{31.60}$ & $\mathbf{32.56}$ & $\mathbf{46.08}$ & $\mathbf{82.72}$ & $\mathbf{0.210}$ & $\mathbf{0.261}$ & $\mathbf{0.317}$ & $\mathbf{0.381}$ & $\mathbf{0.503}$ & $\mathbf{0.564}$ \\
   \end{tabular}}
\label{tab:sr}
\vspace*{-2mm}
\end{table*}

\begin{figure*}[ht!]
\centering
    \begin{subfigure}[t!]{.16\textwidth}
        \centering
        \includegraphics[width=0.1\textwidth]{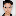}
        %\vspace*{-5mm}
        \caption*{$16^2$}
    \end{subfigure} 
    \begin{subfigure}[t!]{.16\textwidth}
        \includegraphics[width=\textwidth]{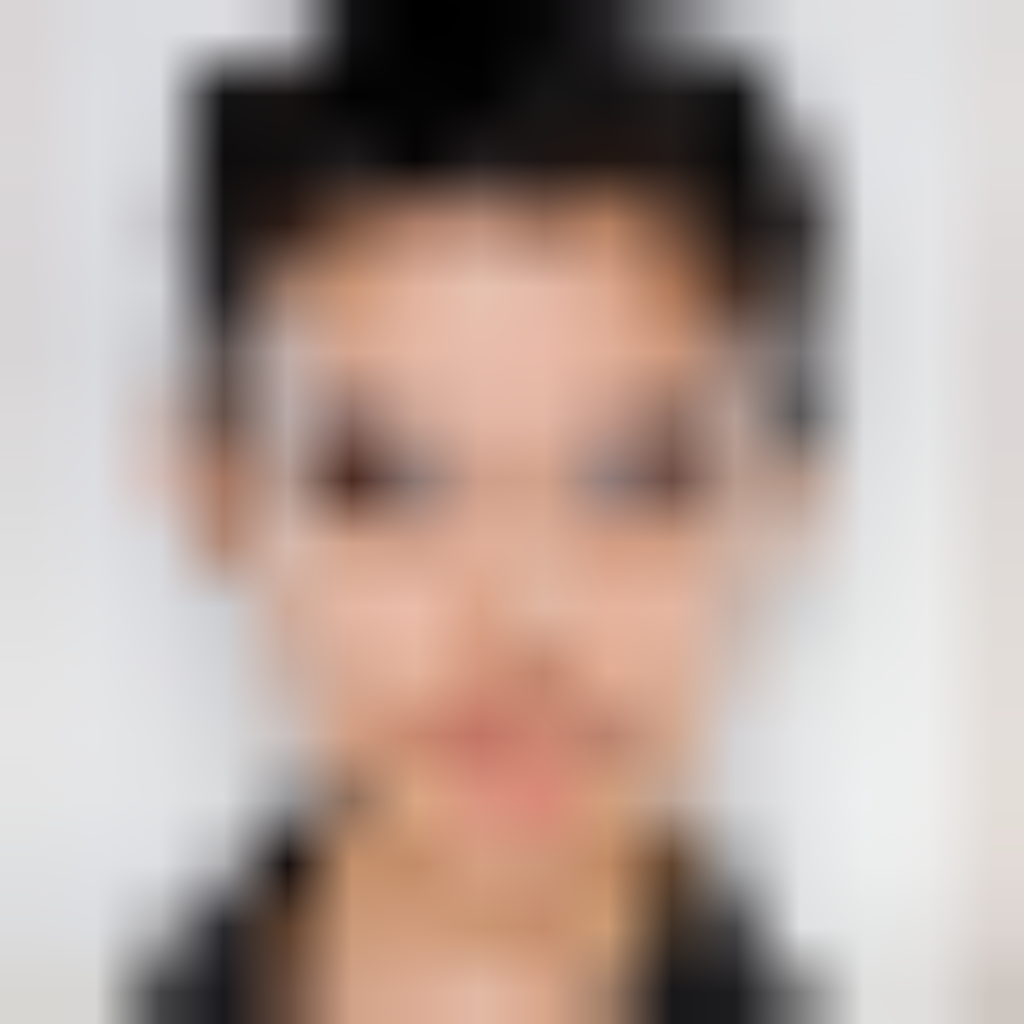}
        \vspace*{-5mm}
        \caption{Bilinear}
    \end{subfigure} 
    \begin{subfigure}[t!]{.16\textwidth}
        \includegraphics[width=\textwidth]{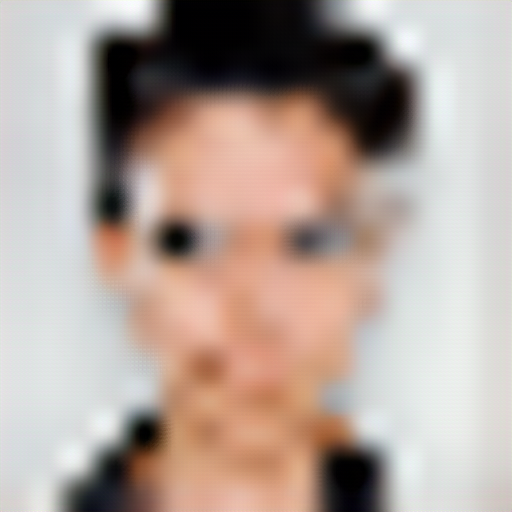}
        \vspace*{-5mm}
        \caption{Super-FAN \protect\cite{Bulat2018SuperFAN}}
    \end{subfigure} 
    \begin{subfigure}[t!]{.16\textwidth}
        \includegraphics[width=\textwidth]{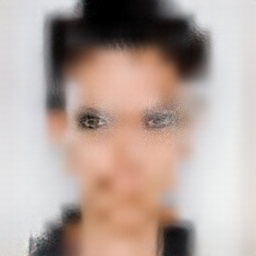}
        \vspace*{-5mm}
        \caption{GWAInet \protect\cite{Dogan2019Exemplar}}
    \end{subfigure} 
    \begin{subfigure}[t!]{.16\textwidth}
        \includegraphics[width=\textwidth]{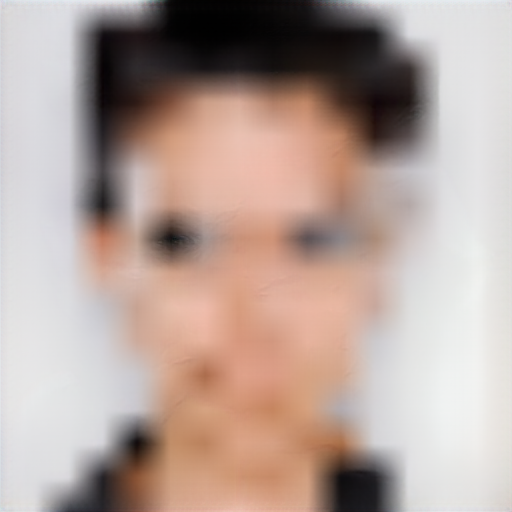}
        \vspace*{-5mm}
        \caption{GFRNet \protect\cite{Li2018GFRNet}}
    \end{subfigure} 
    \begin{subfigure}[t!]{.16\textwidth}
        \includegraphics[width=\textwidth]{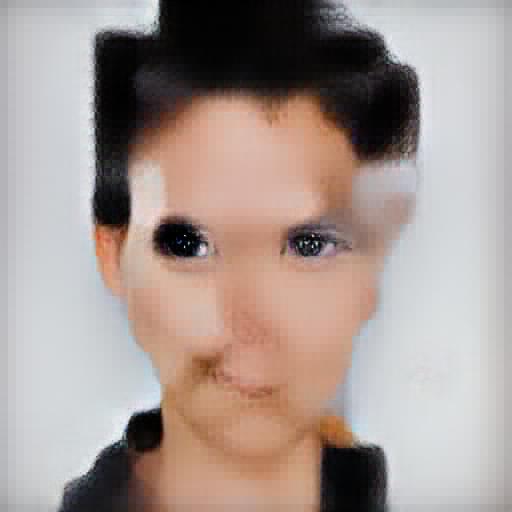}
        \vspace*{-5mm}
        \caption{pix2pixHD \protect\cite{Wang2018Pix2PixHD}}
    \end{subfigure} 
    \\
    \begin{subfigure}[t!]{.16\textwidth}
        \includegraphics[width=\textwidth]{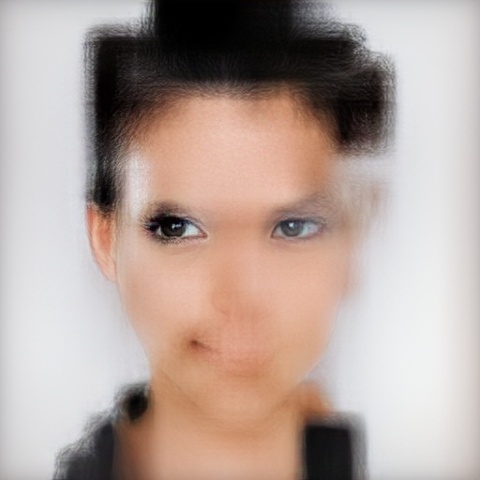}
        \vspace*{-5mm}
        \caption{HiFaceGAN \protect\cite{Yang2020HiFaceGANFR}}
    \end{subfigure} 
    \begin{subfigure}[t!]{.16\textwidth}
        \includegraphics[width=\textwidth]{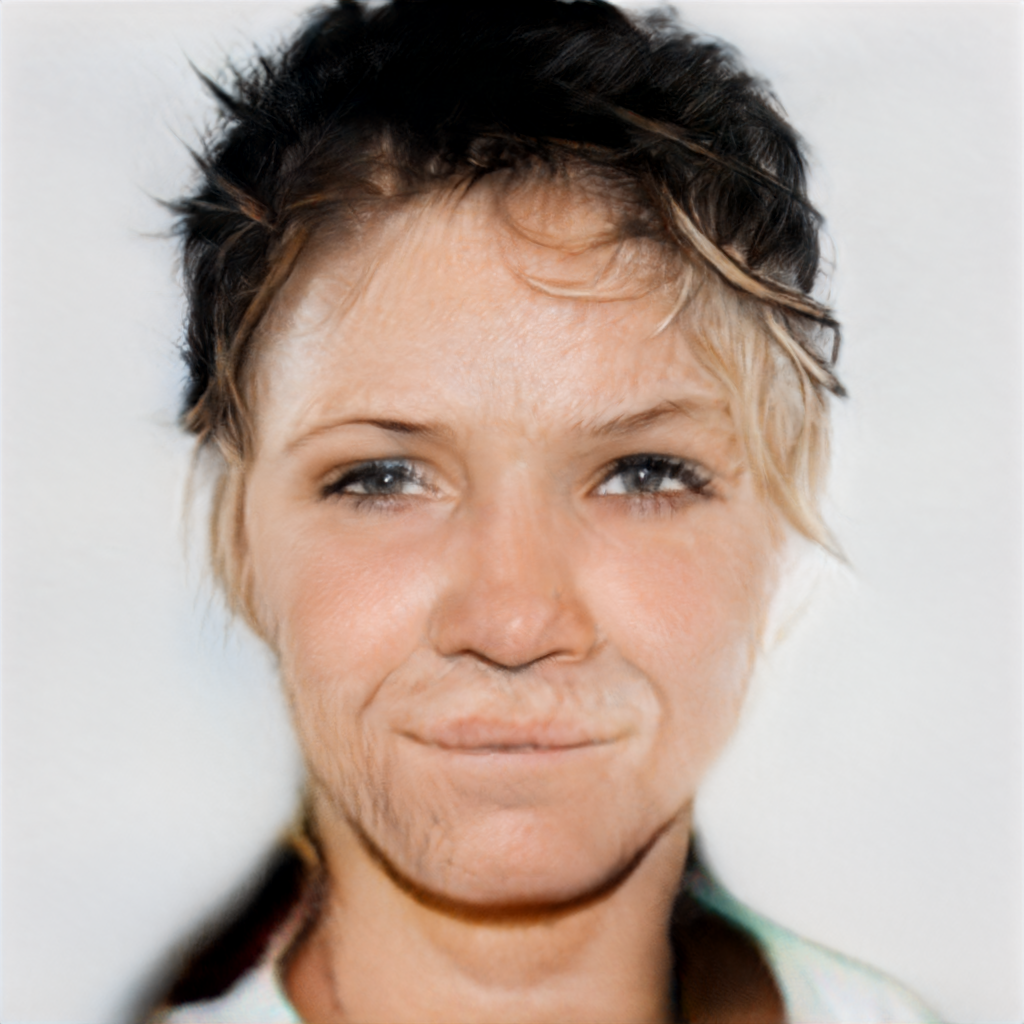}
        \vspace*{-5mm}
        \caption{mGANprior \protect{\cite{Gu2019Prior}}}
    \end{subfigure} 
    \begin{subfigure}[t!]{.16\textwidth}
        \includegraphics[width=\textwidth]{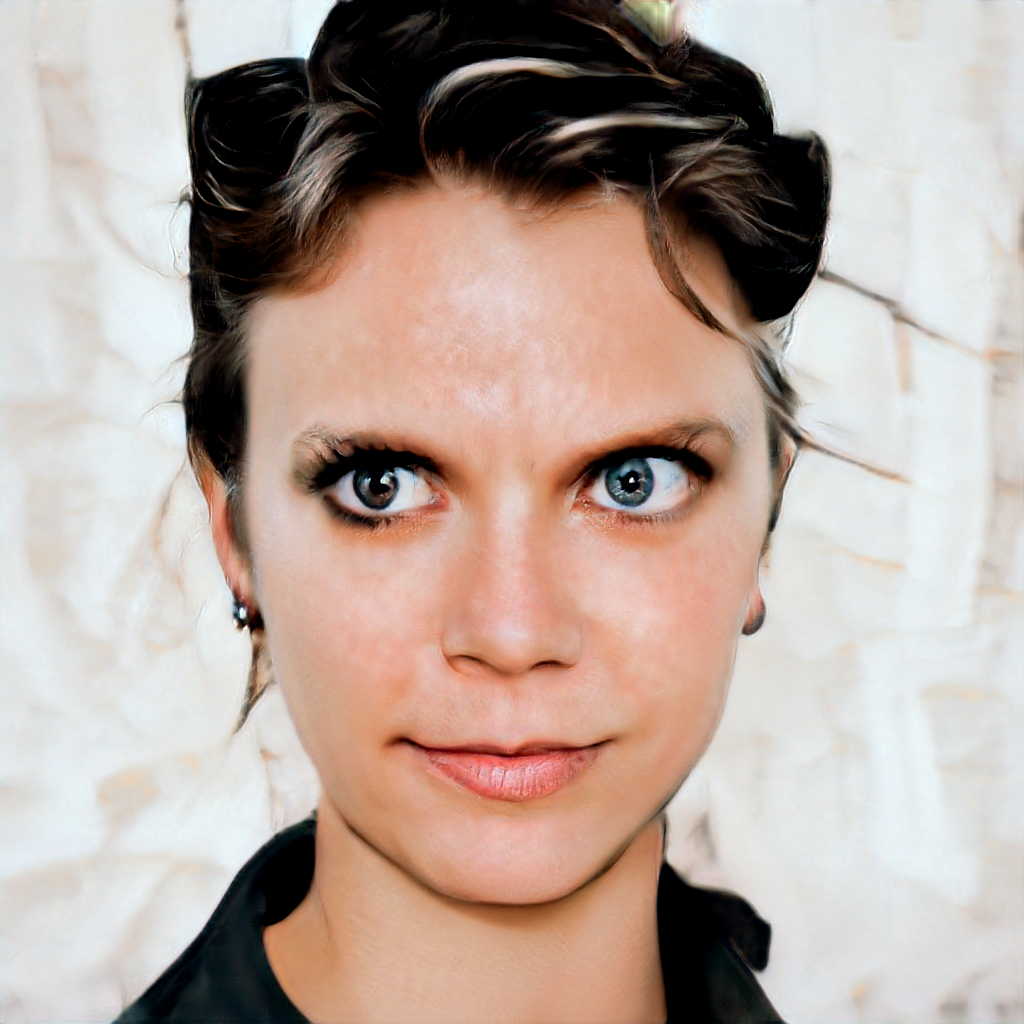}
        \vspace*{-5mm}
        \caption{PULSE \protect{\cite{Menon2020PULSE}}}
    \end{subfigure} 
    \begin{subfigure}[t!]{.16\textwidth}
        \includegraphics[width=\textwidth]{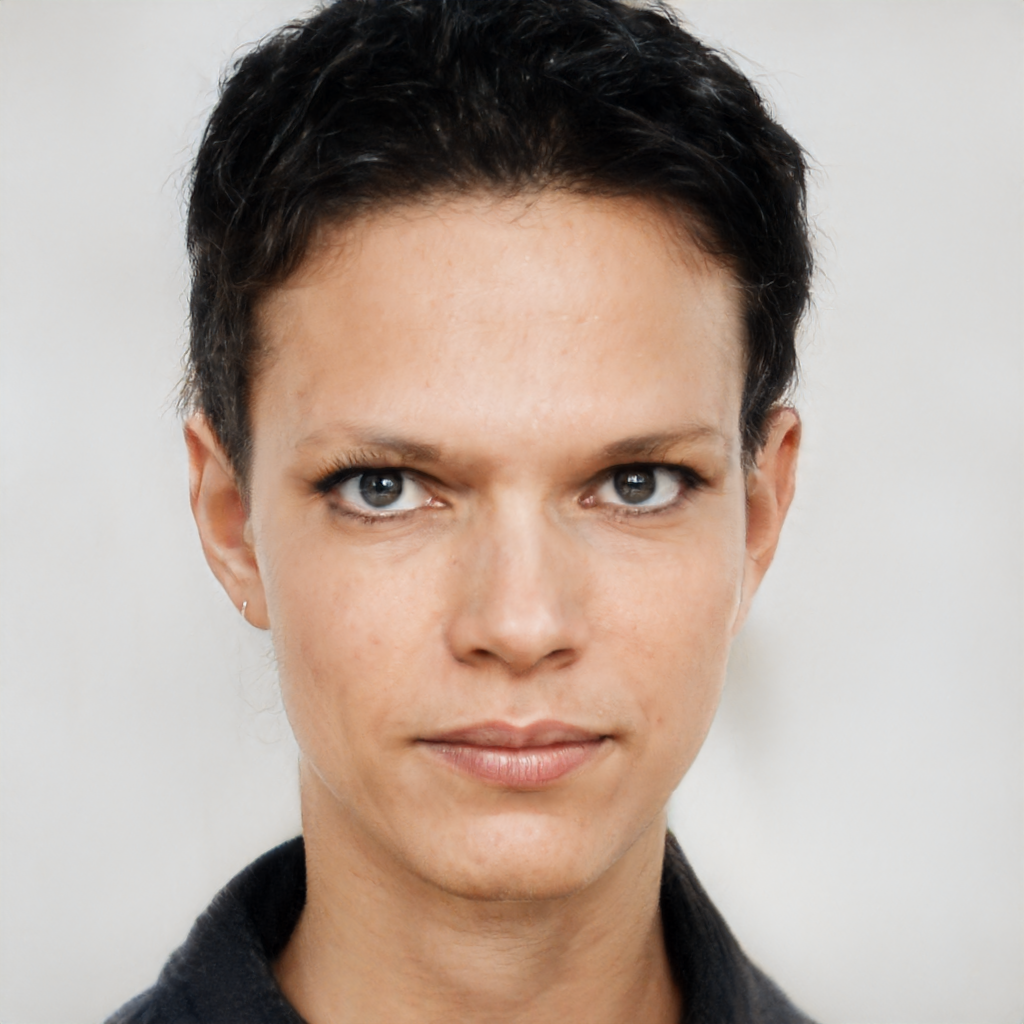}
        \vspace*{-5mm}
        \caption{pSp \protect{\cite{Richardson2020pSp}}}
    \end{subfigure} 
    \begin{subfigure}[t!]{.16\textwidth}
        \includegraphics[width=\textwidth]{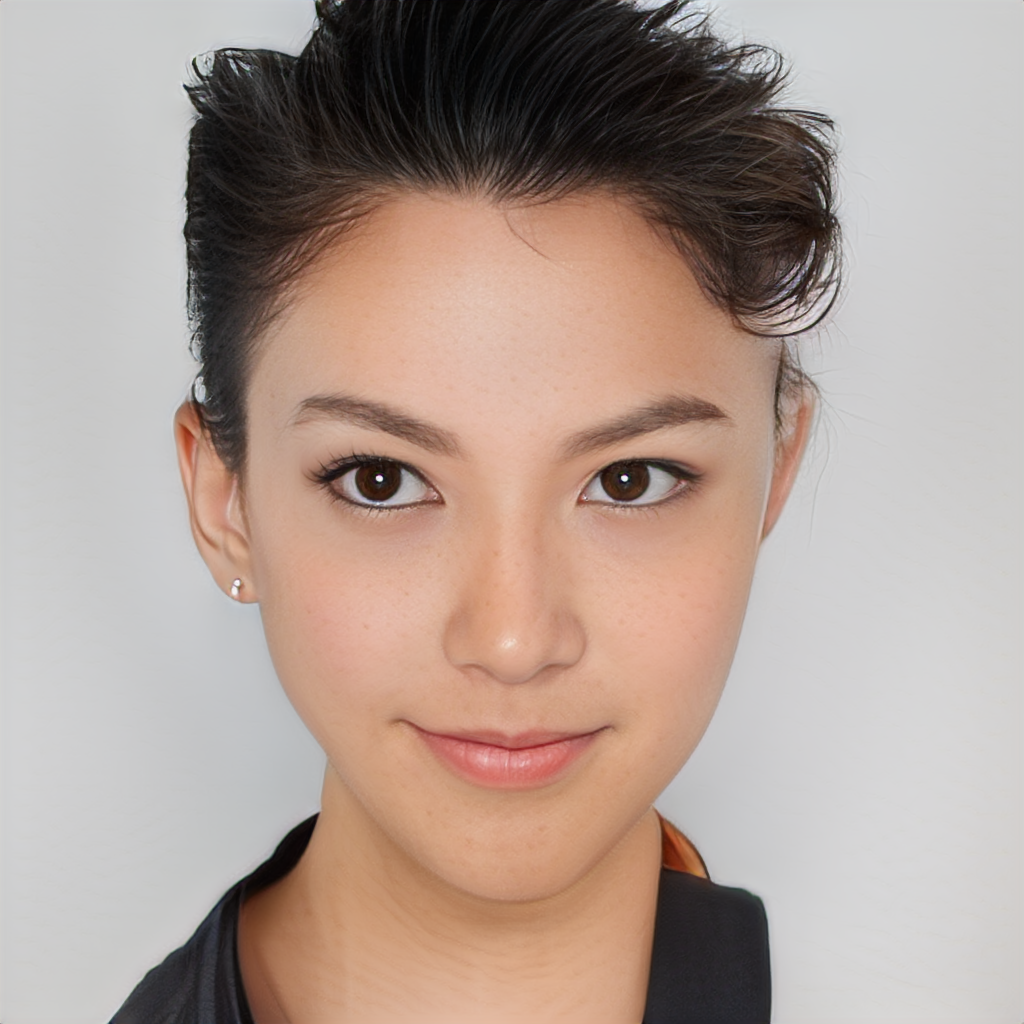}
        \vspace*{-5mm}
        \caption{GPEN}
    \end{subfigure} 
    \begin{subfigure}[t!]{.16\textwidth}
        \includegraphics[width=\textwidth]{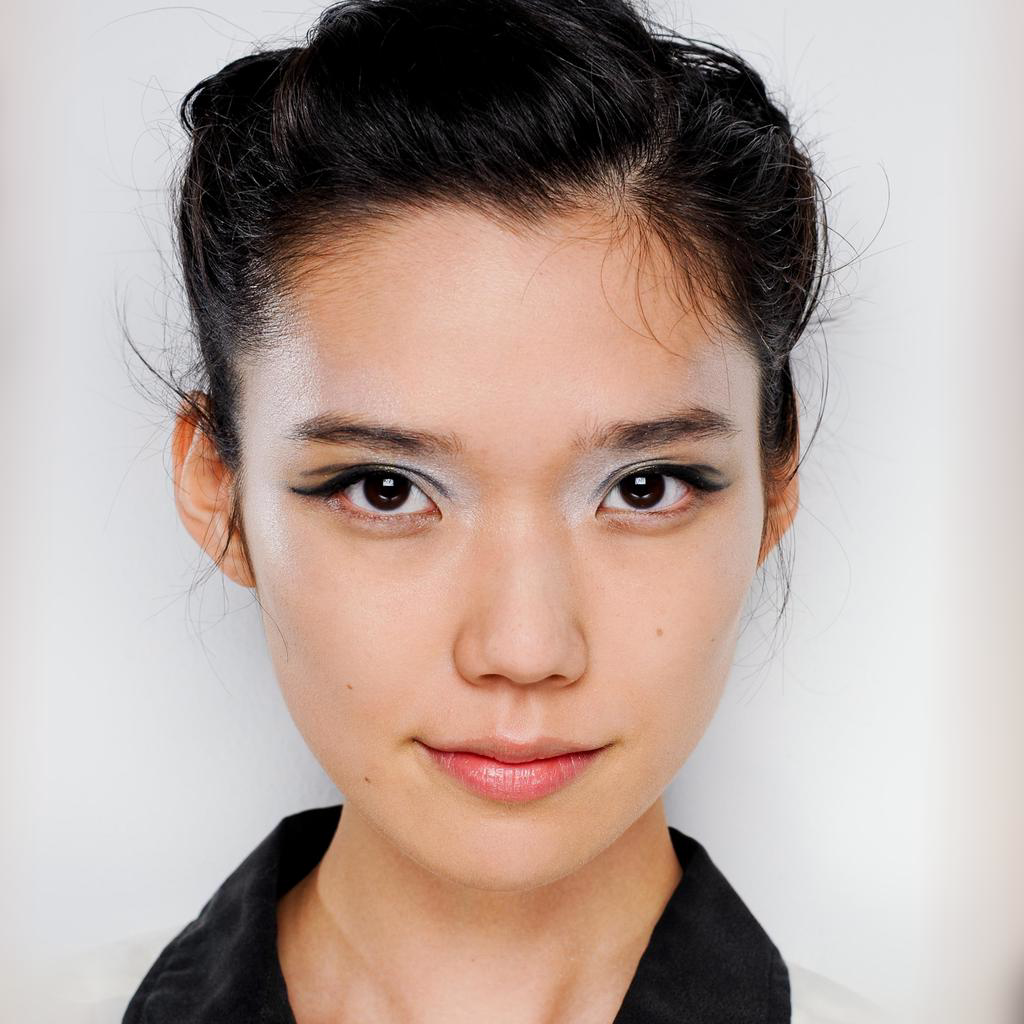}
        \vspace*{-5mm}
        \caption{Ground truth}
    \end{subfigure} 
\vspace*{-1mm}
\caption{Face super-resolution results by state-of-the-art methods. The input image has a resolution of $16^2$.}
\label{fig:fsr16}
\vspace*{-3mm}
\end{figure*}

\subsection{Experiments on Synthetic Images}
To quantitatively compare GPEN with other state-of-the-arts, we first perform experiments on synthetic images. Considering that many face restoration methods \cite{Gu2019Prior,Menon2020PULSE,Richardson2020pSp} are actually designed for FSR instead of BFR, we perform experiments on BFR and FSR separately, where different competing methods are used for fair comparison. 

\textbf{Blind Face Restoration.}
\label{sec:restoration}
By using the degradation model in Eq.~(\ref{eqn:degradation}) and the same set of parameters used in Section~\ref{sec:degradation}, we synthesized a set of LQ face images on the CelebA-HQ dataset for evaluation. We compare GPEN with the latest BFR methods, including Pix2PixHD \cite{Wang2018Pix2PixHD}, Super-FAN \cite{Bulat2018SuperFAN}, GFRNet \cite{Li2018GFRNet}, GWAInet \cite{Dogan2019Exemplar}, DFDNet \cite{Li2020Restore}, HiFaceGAN \cite{Yang2020HiFaceGANFR}. The models trained by the original authors are used in the experiments. We do not compare with those FSR methods \cite{Gu2019Prior,Menon2020PULSE,Richardson2020pSp} in this experiment because they assume a very simple degradation model (e.g., bicubic downsampling) and cannot handle this challenging BFR task. The PSNR, FID and LPIPS results are listed in Table~\ref{tab:restore}. One can see that our GPEN achieves comparable PSNR index to other competing methods, but it achieves significantly better results on FID and LPIPS indices, which are better measures than PSNR for the face image perceptual quality. 

Figure~\ref{fig:comp} compares the BFR results on some degraded face images by the competing methods. One can see that the competing methods fail to produce reasonable face reconstructions. They tend to generate over-smoothed face images with distorted facial structures. However, our GPEN generate visually photo-realistic face images with clear hair, eye, eyebrow, tooth and mustache details. Even the background can also be partially constructed. This clearly validates the advantages of our GPEN model and the training strategy. More visual comparison results can be found in the supplementary file.

\textbf{Face Super-Resolution.}
\label{sec:fsr}
FSR aims to generate an HR image from the input LR version. It can be considered as a special case of BFR, where the image degradation process is specified (i.e., bicubic downsampling). To validate the generality of our GPEN, we still use our model trained for BFR to perform the FSR task, and compare it with those state-of-the-art methods designed for FSR, including Super-FAN \cite{Bulat2018SuperFAN}, GFRNet \cite{Li2018GFRNet}, GWAInet \cite{Dogan2019Exemplar}, DFDNet \cite{Li2020Restore}, HiFaceGAN \cite{Yang2020HiFaceGANFR}, mGANprior \cite{Gu2019Prior}, PULSE \cite{Menon2020PULSE}, and pSp \cite{Richardson2020pSp}. The zooming factor ranges from $8\times$ to $256\times$, and the LR face images are simulated on the CelebA-HQ dataset.

The quantitative results are presented in Table~\ref{tab:sr}. One can see that the naïve bilinear interpolator achieves the best PSNR index, though it cannot restore any facial details. This actually validates that PSNR is not a suitable index to measure FSR quality. GPEN achieves the best FID and LPIPS scores under almost all the zooming factors. Figure~\ref{fig:fsr16} presents a visual comparison example for zooming factor $64\times$. More visual comparison results can be found in the supplementary.

\begin{figure*}[t!]
\centering
    \begin{subfigure}[t!]{.12\textwidth}
        \includegraphics[width=\textwidth]{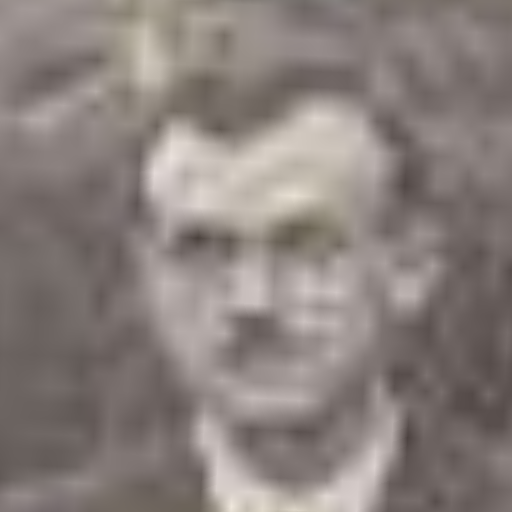}
    \end{subfigure}
    \begin{subfigure}[t!]{.12\textwidth}
        \includegraphics[width=\textwidth]{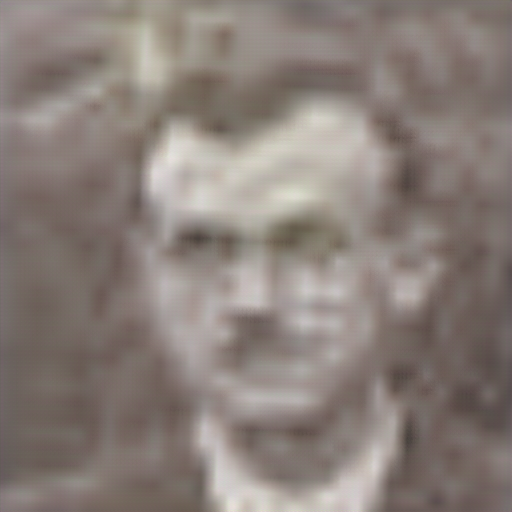}
    \end{subfigure}
    \begin{subfigure}[t!]{.12\textwidth}
        \includegraphics[width=\textwidth]{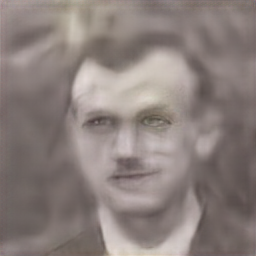}
    \end{subfigure}
    \begin{subfigure}[t!]{.12\textwidth}
        \includegraphics[width=\textwidth]{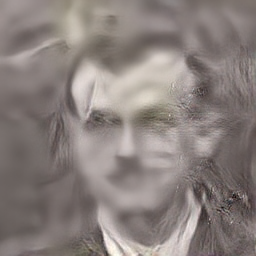}
    \end{subfigure}
    \begin{subfigure}[t!]{.12\textwidth}
        \includegraphics[width=\textwidth]{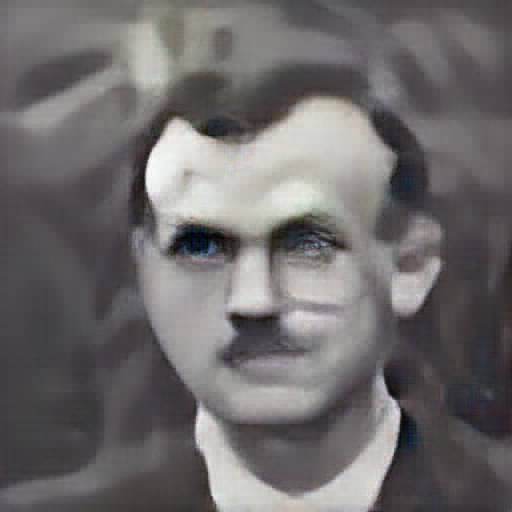}
    \end{subfigure}
    \begin{subfigure}[t!]{.12\textwidth}
        \includegraphics[width=\textwidth]{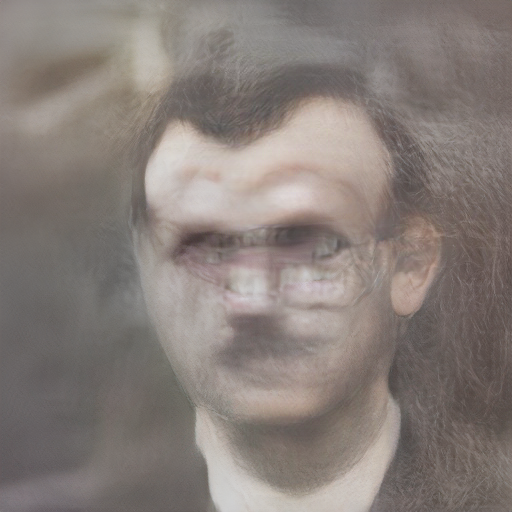}
    \end{subfigure}
    \begin{subfigure}[t!]{.12\textwidth}
        \includegraphics[width=\textwidth]{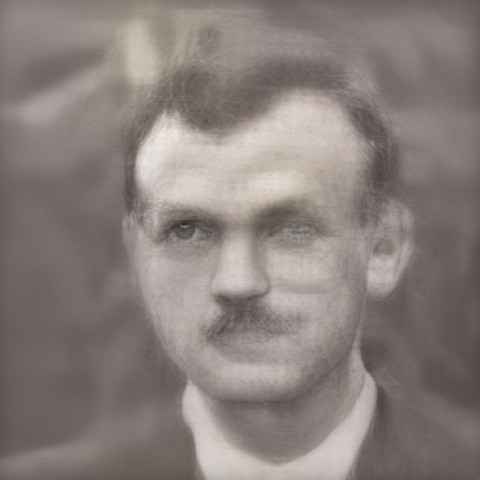}
    \end{subfigure}
    \begin{subfigure}[t!]{.12\textwidth}
        \includegraphics[width=\textwidth]{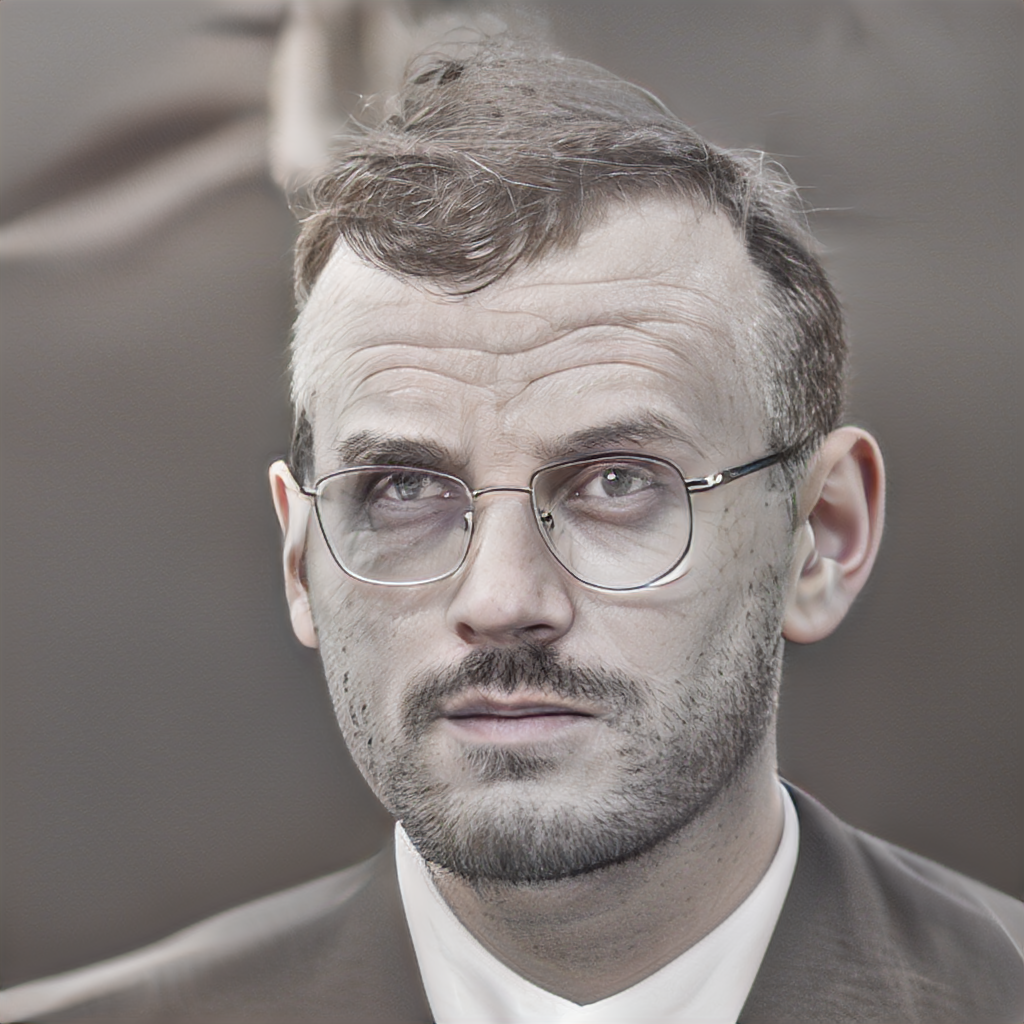}
    \end{subfigure}
    \\
    \begin{subfigure}[t!]{.12\textwidth}
        \includegraphics[width=\textwidth]{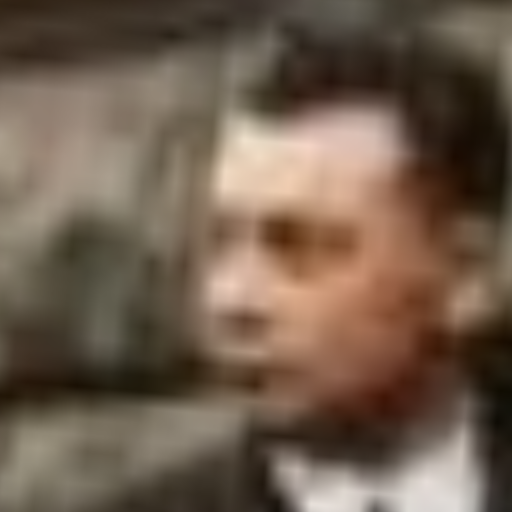}
    \end{subfigure}
    \begin{subfigure}[t!]{.12\textwidth}
        \includegraphics[width=\textwidth]{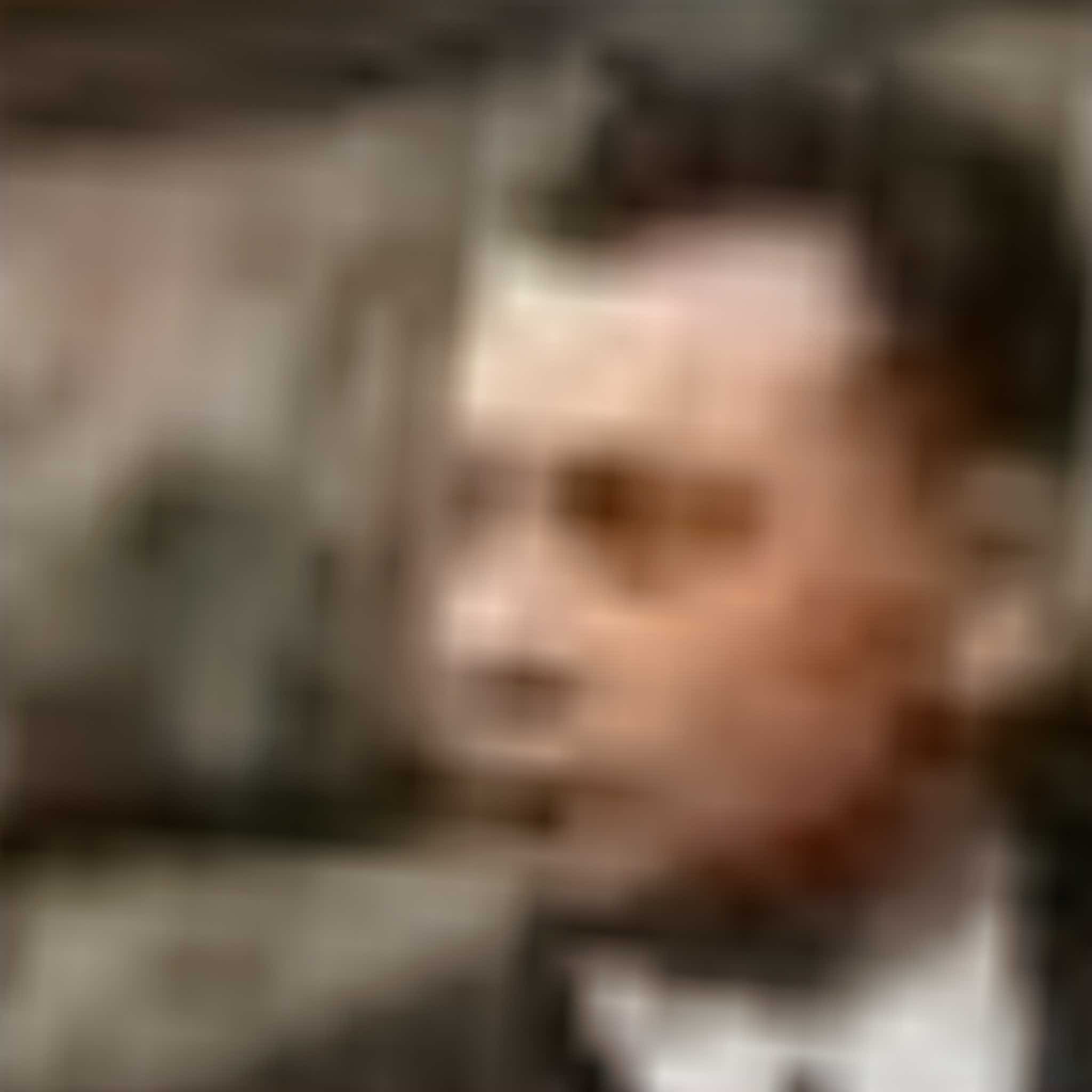}
    \end{subfigure}
    \begin{subfigure}[t!]{.12\textwidth}
        \includegraphics[width=\textwidth]{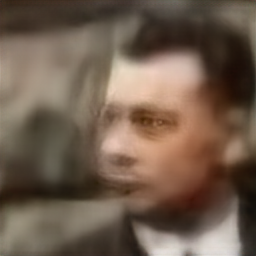}
    \end{subfigure}
    \begin{subfigure}[t!]{.12\textwidth}
        \includegraphics[width=\textwidth]{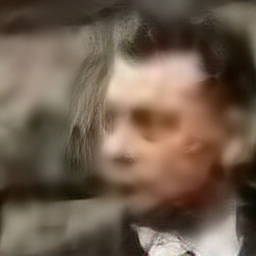}
    \end{subfigure}
    \begin{subfigure}[t!]{.12\textwidth}
        \includegraphics[width=\textwidth]{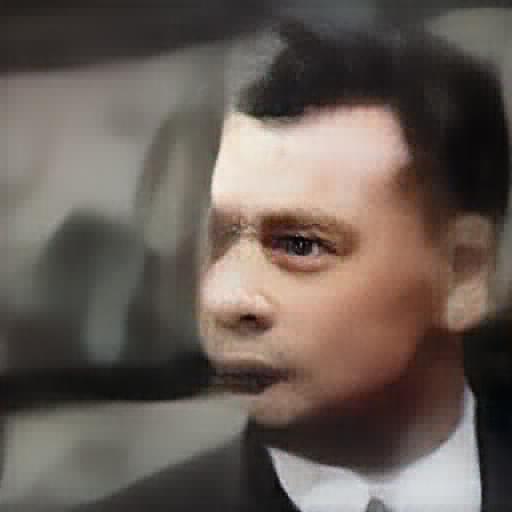}
    \end{subfigure}
    \begin{subfigure}[t!]{.12\textwidth}
        \includegraphics[width=\textwidth]{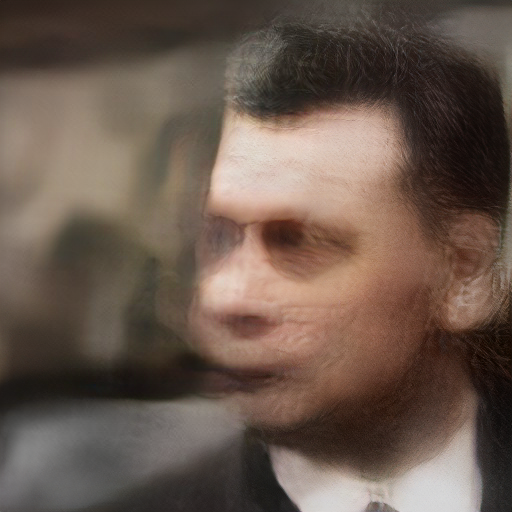}
    \end{subfigure}
    \begin{subfigure}[t!]{.12\textwidth}
        \includegraphics[width=\textwidth]{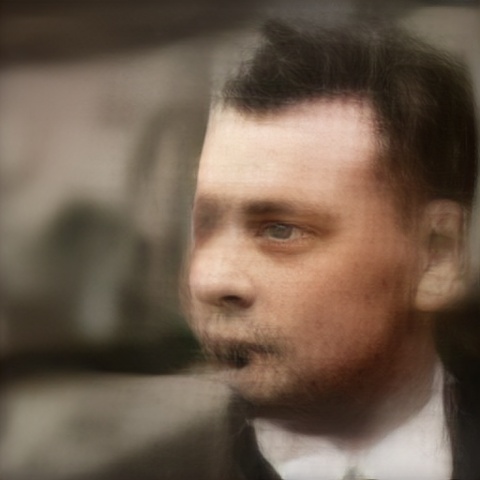}
    \end{subfigure}
    \begin{subfigure}[t!]{.12\textwidth}
        \includegraphics[width=\textwidth]{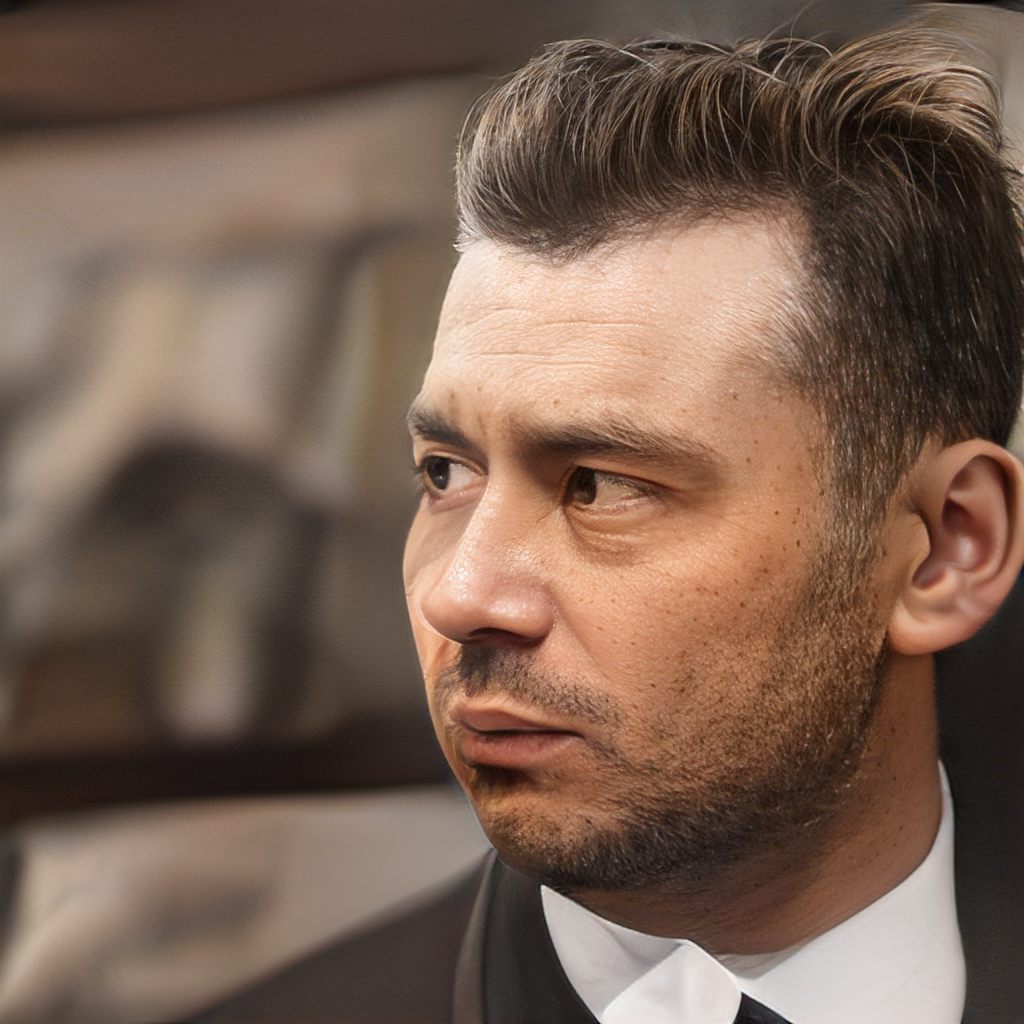}
    \end{subfigure}
    \\
    \begin{subfigure}[t!]{.12\textwidth}
        \includegraphics[width=\textwidth]{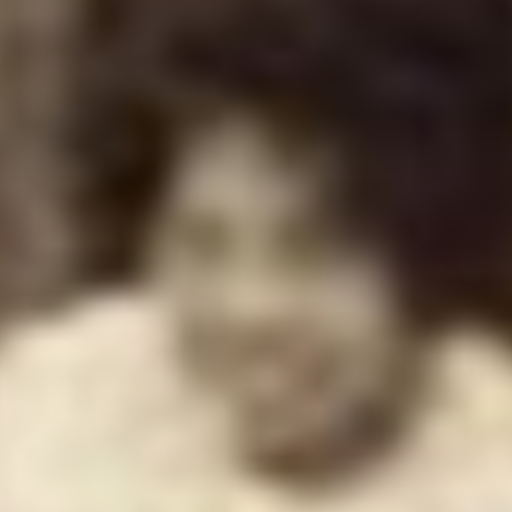}
        \vspace*{-5mm}
        \caption{}
    \end{subfigure}
    \begin{subfigure}[t!]{.12\textwidth}
        \includegraphics[width=\textwidth]{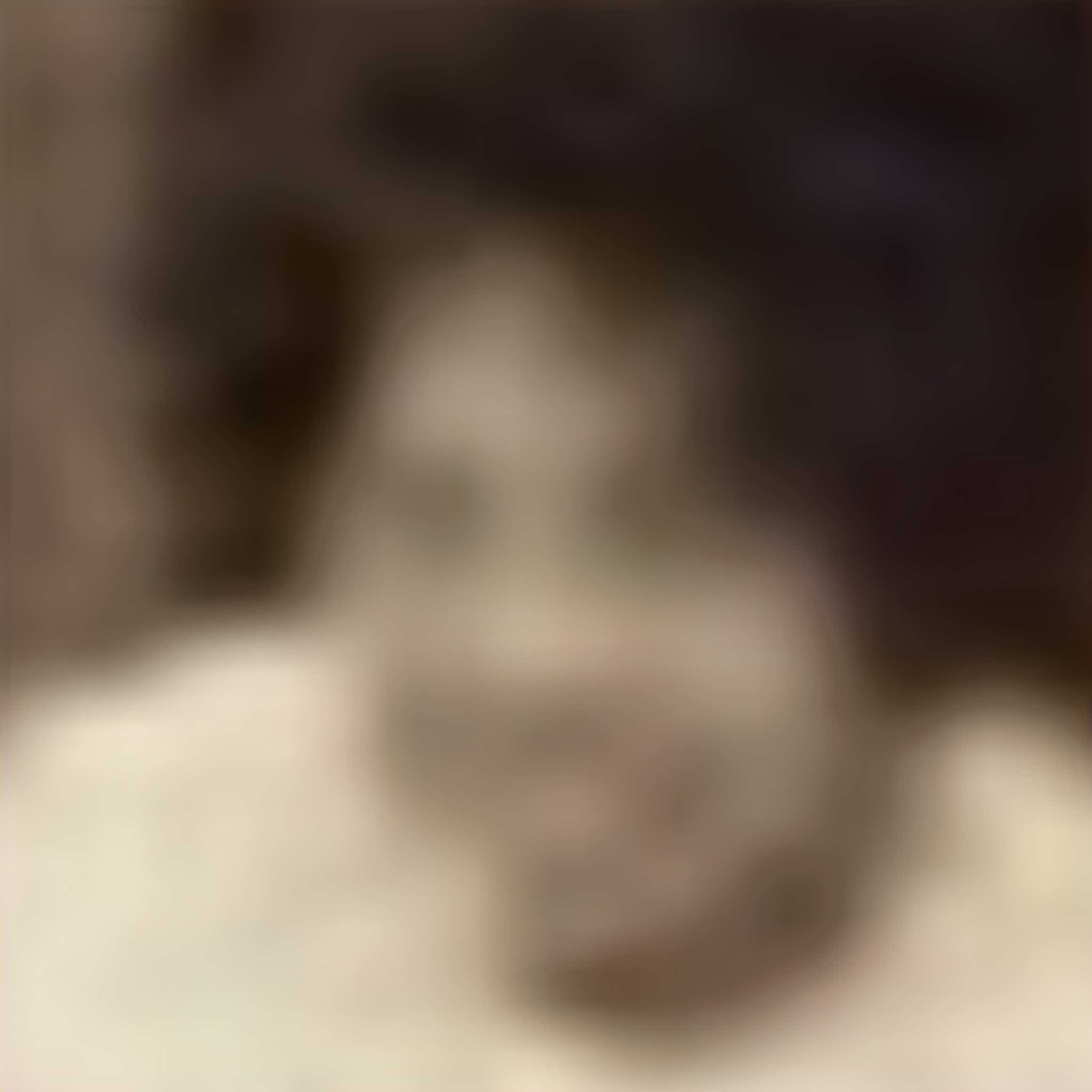}
        \vspace*{-5mm}
        \caption{}
    \end{subfigure}
    \begin{subfigure}[t!]{.12\textwidth}
        \includegraphics[width=\textwidth]{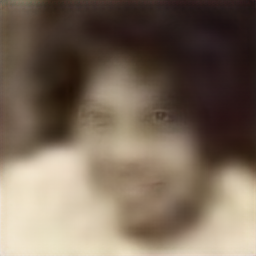}
        \vspace*{-5mm}
        \caption{}
    \end{subfigure}
    \begin{subfigure}[t!]{.12\textwidth}
        \includegraphics[width=\textwidth]{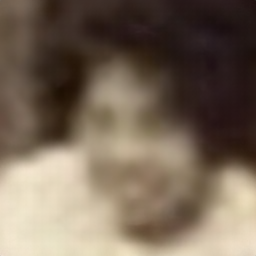}
        \vspace*{-5mm}
        \caption{}
    \end{subfigure}
    \begin{subfigure}[t!]{.12\textwidth}
        \includegraphics[width=\textwidth]{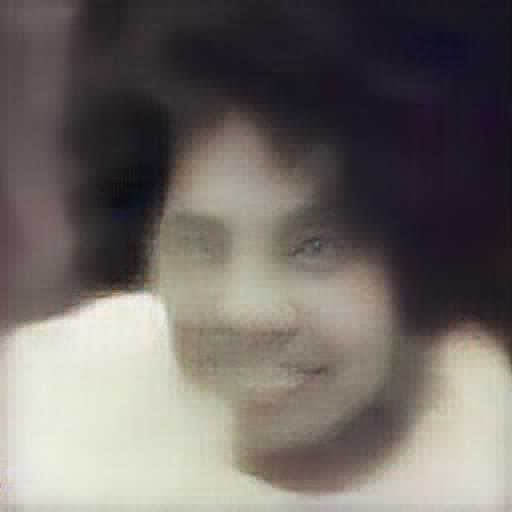}
        \vspace*{-5mm}
        \caption{}
    \end{subfigure}
    \begin{subfigure}[t!]{.12\textwidth}
        \includegraphics[width=\textwidth]{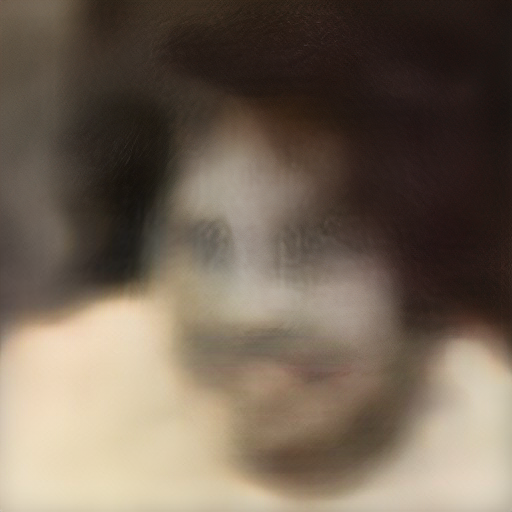}
        \vspace*{-5mm}
        \caption{}
    \end{subfigure}
    \begin{subfigure}[t!]{.12\textwidth}
        \includegraphics[width=\textwidth]{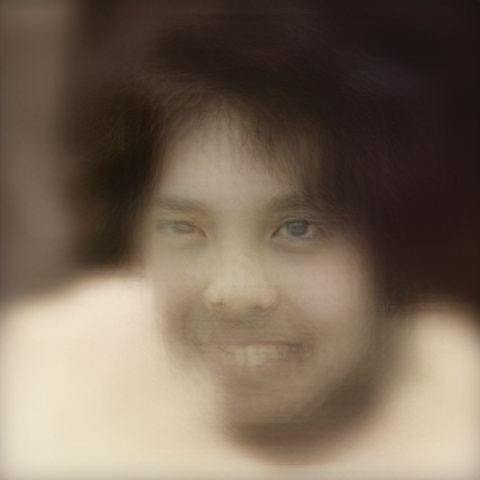}
        \vspace*{-5mm}
        \caption{}
    \end{subfigure}
    \begin{subfigure}[t!]{.12\textwidth}
        \includegraphics[width=\textwidth]{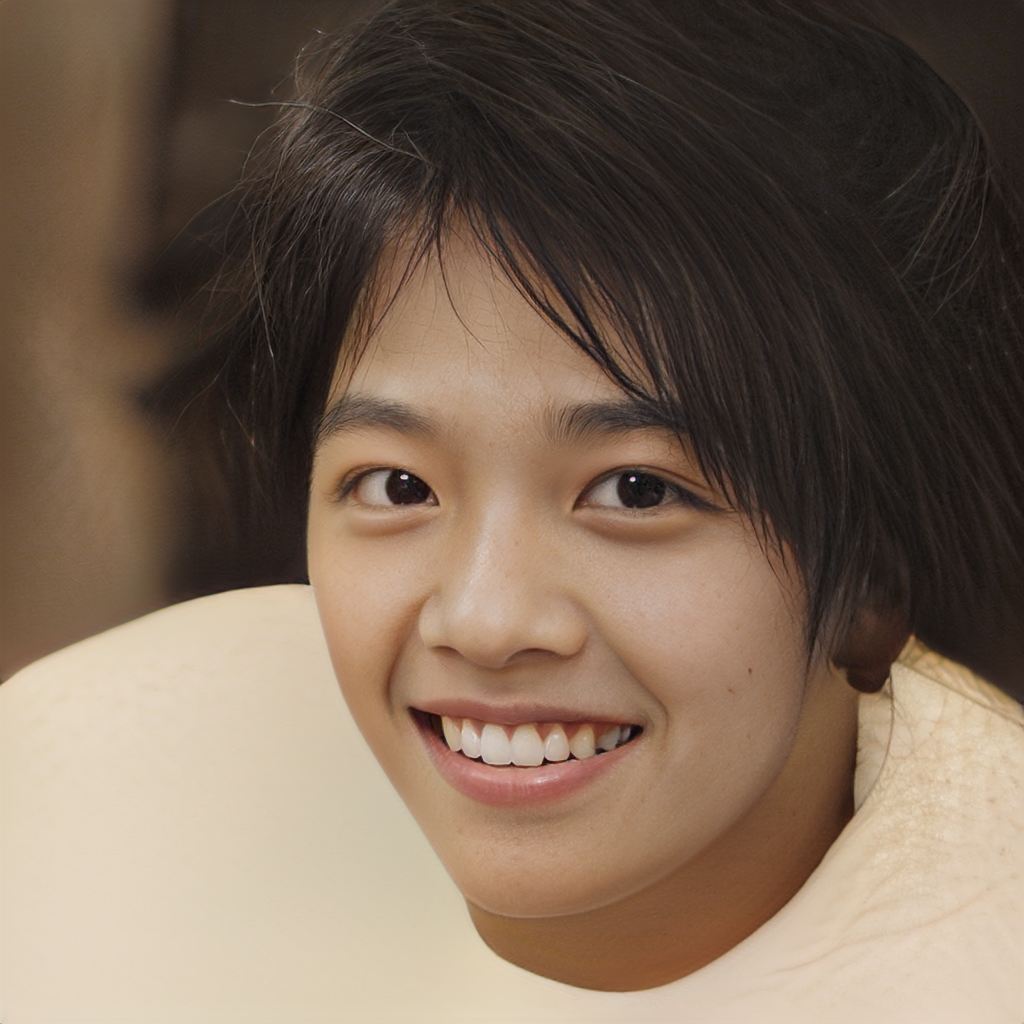}
        \vspace*{-5mm}
        \caption{}
    \end{subfigure}
\vspace*{-5mm}
\caption{Blind face restoration results on real degraded faces in the wild. (a) Real degraded faces; (b) Super-FAN \protect{\cite{Bulat2018SuperFAN}}; (c) GFRNet \protect{\cite{Li2018GFRNet}}; (d) GWAInet \protect{\cite{Dogan2019Exemplar}}; (e) Pix2PixHD \protect{\cite{Wang2018Pix2PixHD}}; (f) DFDNet \protect{\cite{Li2020Restore}}; (g) HiFaceGAN \protect{\cite{Yang2020HiFaceGANFR}}; (h) GPEN.}
\label{fig:realcomp}
\vspace*{-3mm}
\end{figure*} 

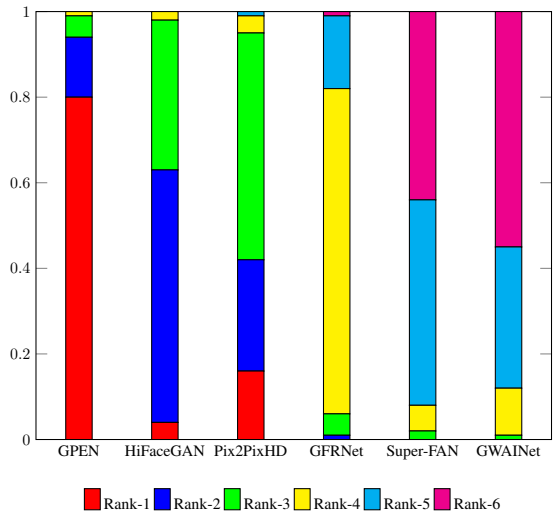
\begin{figure}
\centering
\vspace*{0mm}\begin{tikzpicture}[thick,scale=1, every node/.style={scale=0.6}]
   \begin{axis}[
      ybar stacked, ymin=0, ymax=1, 
      symbolic x coords={GPEN, HiFaceGAN, Pix2PixHD, GFRNet, Super-FAN, GWAINet},
      xtick=data,
      %x tick label style={rotate=45,anchor=east},
      legend style={draw=none,at={(0.5,-0.1)},anchor=north,legend columns=-1},
      ]

   \addplot [fill=red] coordinates
      {(GPEN,0.8) (HiFaceGAN,0.04) (Pix2PixHD,0.16) (GFRNet,0) (Super-FAN,0) (GWAINet,0)};
   \addplot [fill=blue] coordinates
      {(GPEN,0.14) (HiFaceGAN,0.59) (Pix2PixHD,0.26) (GFRNet,0.01) (Super-FAN,0) (GWAINet,0)};
   \addplot [fill=green] coordinates
      {(GPEN,0.05) (HiFaceGAN,0.35) (Pix2PixHD,0.53) (GFRNet,0.05) (Super-FAN,0.02) (GWAINet,0.01)};
   \addplot [fill=yellow] coordinates
      {(GPEN,0.01) (HiFaceGAN,0.02) (Pix2PixHD,0.04) (GFRNet,0.76) (Super-FAN,0.06) (GWAINet,0.11)};
   \addplot [fill=cyan] coordinates
      {(GPEN,0) (HiFaceGAN,0) (Pix2PixHD,0.01) (GFRNet,0.17) (Super-FAN,0.48) (GWAINet,0.33)};
   \addplot [fill=magenta] coordinates
      {(GPEN,0) (HiFaceGAN,0) (Pix2PixHD,0) (GFRNet,0.01) (Super-FAN,0.44) (GWAINet,0.55)};
   \legend{Rank-1, Rank-2, Rank-3, Rank-4, Rank-5, Rank-6}
   \end{axis}
\end{tikzpicture}
\caption{User study results of different BFR methods.}
\label{fig:ustudy}
\vspace*{-5mm}
\end{figure}

\subsection{Experiments on Images in the Wild}
Finally, we perform experiments on real-world LQ face images, which suffer from complex unknown degradations. We collected $1,000$ LQ face images from internet for testing. The BFR methods Pix2PixHD \cite{Wang2018Pix2PixHD}, Super-FAN \cite{Bulat2018SuperFAN}, GFRNet \cite{Li2018GFRNet}, GWAInet \cite{Dogan2019Exemplar}, DFDNet \cite{Li2020Restore} and HiFaceGAN \cite{Yang2020HiFaceGANFR} are used in the comparison. Figure~\ref{fig:realcomp} shows the BFR results on three images. One can see that the competing methods fail to restore the facial details. This is mainly because they are trained on synthesized data but have limited generalization capability to the images in the wild. Our method manages to overcome this difficulty by the carefully designed GAN prior embedding and fine-tuning strategies. It not only preserves well the global structure of the face, but also generates realistic details on the face components (e.g., hair, eye, mouth, etc.). Our GPEN can also be successfully used to renovate old photos, as we demonstrated in Figure~\ref{fig:oldphotos}. Please refer to the supplementary material for more results.

%\textbf{User Study.} Commonly used quantitative metrics like PSNR and SSIM do not strongly correlate with human visual perception to image quality. Therefore, we conduct a user survey as a subjective assessment to analyze the performance of our method in comparison to competing methods. The restored results of our GPEN method, Pix2PixHD \cite{Wang2018Pix2PixHD}, Super-FAN \cite{Bulat2018SuperFAN}, GFRNet \cite{Li2018GFRNet}, GWAInet \cite{Dogan2019Exemplar}, DFDNet \cite{Li2020Restore} and HiFaceGAN \cite{Yang2020HiFaceGANFR} on $113$ real LQ face images collected from internet are presented in a randomized sequence. Human raters are asked to rank the six versions according to the perceptual quality. Finally, we collect $1915$ votes from $17$ human raters. The summarized results are presented in Figure~\ref{fig:ustudy}. As shown, our GPEN method gets much more votes of rank-1 than the state-of-the arts.
Since the commonly used quantitative metrics like PSNR and SSIM do not strongly correlate with human visual perception to image quality, we conduct a user study as a subjective assessment on the performance of our method and the competing methods. The BFR results of GPEN, Pix2PixHD \cite{Wang2018Pix2PixHD}, Super-FAN \cite{Bulat2018SuperFAN}, GFRNet \cite{Li2018GFRNet}, GWAInet \cite{Dogan2019Exemplar}, DFDNet \cite{Li2020Restore} and HiFaceGAN \cite{Yang2020HiFaceGANFR} on $113$ real-world LQ face images collected from internet are presented in a random order to $17$ volunteers for subjective evaluation. The volunteers are asked to rank the six BFR outputs of each input image according to their perceptual quality. Finally, we collect $1,915$ votes, and the statistics are presented in Figure~\ref{fig:ustudy}. As can be seen, our GPEN method receives much more rank-1 votes than the other state-of-the arts.

%\section{Discussion}
%We have provided a novel insight on how to make better use of the GAN priors. Previous methods mainly keep the GANs unstained and work on optimizing latent space. We are the first to use the GAN priors in a transfer learning way. Moreover, both the generative and discriminative models are considered as the GAN priors in our paper, while previous works fail to do so. Both the generator and discriminator matter significantly. In our experiments, the discriminator obtained by the end-to-end training is incapable of telling apart faces with/widthout detailed features due to the global-first theory, leading to over-smoothing results. And the generator is the key to provide the framework the capability of generating high-quality faces. If the generator is being trained from scratch, the framework can hardly reach the point of generating faces with faithful quality. Our experiments on extreme super-resolution provde it.

\section{Conclusion and Discussion}
We proposed a simple yet effective GAN prior embedded network, namely GPEN, for BFR in the wild. By embedding a pre-trained GAN into a U-shaped DNN as a decoder, and fine-tuning the whole network with artificially degraded face images, our model learned to generate high quality face images from severely degraded ones. Our extensive experiments on synthetic data and real-world images demonstrated that GPEN outperforms the latest state-of-the-arts significantly, restoring clear facial details while retaining properly the image background. The proposed method can also be applied to other tasks such as face inpainting and face colorization. Some preliminary results were provided in the supplementary material. 

%The proposed GPEN does not allow multiple HQ images to be generated from a single LQ image in the current form. StyleGAN controls the synthesis via style mixing. However, this operation would lead to inconsistent image background in GPEN. In the future, we will make GPEN allow for diversity by taking an extra HQ face image as input to control the idendity due to preserving the identity of a severely degraded face is incredibly hard and even somewhat impossible.
The proposed GPEN does not allow multiple HQ images to be generated from a single LQ image in its current form. StyleGAN controls the synthesis via style mixing; however, such an operation may lead to inconsistent image background in GPEN. In the future, we will extend GPEN to allow multiple HQ outputs for a given LQ image. For example, we can use an extra HQ face image as a reference so that different HQ outputs can be generated by GPEN for different reference images.

{\small
\bibliographystyle{ieee_fullname}
\bibliography{egbib}

\begin{thebibliography}{10}\itemsep=-1pt

\bibitem{Abdal2019Img2StyleGAN}
Rameen Abdal, Yipeng Qin, and Peter Wonka.
\newblock Image2stylegan: How to embed images into the stylegan latent space?
\newblock In {\em ICCV}, 2019.

\bibitem{Bakerand2000Hallucinate}
S. Bakerand and T. Kanade.
\newblock Hallucinating faces.
\newblock In {\em IEEE International Conference on Automatic Face and Gesture
  Recognition}, 2000.

\bibitem{Bourlai2011Restore}
Thirimachos Bourlai, Arun Ross, and Anil~K. Jain.
\newblock Restoring degraded face images: A case study in matching faxed,
  printed, and scanned photos.
\newblock {\em IEEE Transactions on Information Forensics and Security},
  6(2):371--384, 2011.

\bibitem{Brock2019BigGAN}
Andrew Brock, Jeff Donahue, and Karen Simonyan.
\newblock Large scale {GAN} training for high fidelity natural image synthesis.
\newblock In {\em ICLR}, 2019.

\bibitem{Bulat2018SuperFAN}
Adrian Bulat and Georgios Tzimiropoulos.
\newblock Super-fan: Integrated facial landmark localization and
  super-resolution of real-world low resolution faces in arbitrary poses with
  gans.
\newblock In {\em CVPR}, 2018.

\bibitem{Chen2005Global}
Lin Chen.
\newblock The topological approach to perceptual organization.
\newblock {\em Visual Cognition}, 12(4):553–637, 2005.

\bibitem{Chen2018FSRNet}
Yu Chen, Ying Tai, Xiaoming Liu, Chunhua Shen, and Jian Yang.
\newblock Fsrnet: End-to-end learning face super-resolution with facial priors.
\newblock In {\em CVPR}, 2018.

\bibitem{Choi2018Stargan}
Yunjey Choi, Minje Choi, Munyoung Kim, Jung-Woo Ha, Sunghun Kim, and Jaegul
  Choo.
\newblock Stargan: Unified generative adversarial networks for multi-domain
  image-to-image translation.
\newblock In {\em CVPR}, 2018.

\bibitem{Dogan2019Exemplar}
Berk Dogan, Shuhang Gu, and Radu Timofte.
\newblock Exemplar guided face image super-resolution without facial landmarks.
\newblock In {\em CVPRW}, 2019.

\bibitem{Frgier2019Mind2MindT}
Ya{\"e}l Fr{\'e}gier and Jean-Baptiste Gouray.
\newblock Mind2mind : transfer learning for gans.
\newblock {\em arXiv}, 2019.

\bibitem{Goodfellow2014GAN}
Ian~J. Goodfellow, Jean Pouget-Abadie, Mehdi Mirza, Bing Xu, David
  Warde-Farley, Sherjil Ozair, Aaron Courville, and Yoshua Bengio.
\newblock Generative adversarial nets.
\newblock In {\em NIPS}, page 2672–2680, 2014.

\bibitem{Gu2019Prior}
Jinjin Gu, Yujun Shen, and Bolei Zhou.
\newblock Image processing using multi-code gan prior.
\newblock {\em ArXiv}, 2019.

\bibitem{Guo2019CBDNet}
Shi Guo, Zifei Yan, Kai Zhang, Wangmeng Zuo, and Lei Zhang.
\newblock Toward convolutional blind denoising of real photographs.
\newblock {\em CVPR}, 2019.

\bibitem{Hand2018Phase}
Paul Hand, Oscar Leong, and Vladislav Voroninski.
\newblock Phase retrieval under a generative prior.
\newblock In {\em NIPS}, 2018.

\bibitem{Heusel2017FID}
Martin Heusel, Hubert Ramsauer, Thomas Unterthiner, Bernhard Nessler, and Sepp
  Hochreiter.
\newblock Gans trained by a two time-scale update rule converge to a local nash
  equilibrium.
\newblock In {\em NIPS}, 2017.

\bibitem{Hu20203dprior}
Xiaobin Hu, Wenqi Ren, John Lamaster, Xiaochun Cao, Xiaoming Li, Zechao Li,
  Bjoern Menze, and Wei Liu.
\newblock Face super-resolution guided by 3d facial priors.
\newblock In {\em ECCV}, 2020.

\bibitem{Huang2017Wavelet}
Huaibo Huang, Ran He, Zhenan Sun, and Tieniu Tan.
\newblock Wavelet-srnet: A wavelet-based cnn for multi-scale face super
  resolution.
\newblock In {\em ICCV}, 2017.

\bibitem{Isola2017Pix2Pix}
Phillip Isola, Jun-Yan Zhu, Tinghui Zhou, and Alexei~A Efros.
\newblock Image-to-image translation with conditional adversarial networks.
\newblock {\em CVPR}, 2017.

\bibitem{Johnson2016Perceptual}
Justin Johnson, Alexandre Alahi, and Li Fei-Fei.
\newblock Perceptual losses for real-time style transfer and super-resolution.
\newblock In {\em ECCV}, 2016.

\bibitem{Karras2018PGGAN}
Tero Karras, Timo Aila, and Samuli Laine.
\newblock Progressive growing of gans for improved quality, stability, and
  variation.
\newblock In {\em ICLR}, 2018.

\bibitem{Karras2018StyleGAN}
Tero Karras, Samuli Laine, and Timo Aila.
\newblock A style-based generator architecture for generative adversarial
  networks.
\newblock {\em ArXiv}, 2018.

\bibitem{Karras2019StyleGAN2}
Tero Karras, Samuli Laine, Miika Aittala, Janne Hellsten, Jaakko Lehtinen, and
  Timo Aila.
\newblock Analyzing and improving the image quality of stylegan.
\newblock {\em ArXiv}, 2019.

\bibitem{Kim2019PFSR}
Deokyun Kim, Minseon Kim, Gihyun Kwon, and Dae-Shik Kim.
\newblock Progressive face super-resolution via attention to facial landmark.
\newblock {\em ArXiv}, 2019.

\bibitem{Kupyn2017DeblurGAN}
Orest Kupyn, Volodymyr Budzan, Mykola Mykhailych, Dmytro Mishkin, and Jiri
  Matas.
\newblock Deblurgan: Blind motion deblurring using conditional adversarial
  networks.
\newblock {\em ArXiv}, 2017.

\bibitem{Ledig2017SRGAN}
Christian Ledig, Lucas Theis, Ferenc Huszar, Jose Caballero, Andrew Cunningham,
  Alejandro Acosta, Andrew Aitken, Alykhan Tejani, Johannes Totz, Zehan Wang,
  and Wenzhe Shi.
\newblock Photo-realistic single image super-resolution using a generative
  adversarial network.
\newblock In {\em CVPR}, 2017.

\bibitem{Lee2018DRIT}
Hsin-Ying Lee, Hung-Yu Tseng, Jia-Bin Huang, Maneesh~Kumar Singh, and
  Ming-Hsuan Yang.
\newblock Diverse image-to-image translation via disentangled representations.
\newblock In {\em ECCV}, 2018.

\bibitem{Li2020Restore}
Xiaoming Li, Chaofeng Chen, Shangchen Zhou, Xianhui Lin, Wangmeng Zuo, and Lei
  Zhang.
\newblock Blind face restoration via deep multi-scale component dictionaries.
\newblock In {\em ECCV}, 2020.

\bibitem{Li2020ASFFNet}
Xiaoming Li, Wenyu Li, Dongwei Ren, Hongzhi Zhang, Meng Wang, and Wangmeng Zuo.
\newblock Enhanced blind face restoration with multi-exemplar images and
  adaptive spatial feature fusion.
\newblock In {\em CVPR}, 2020.

\bibitem{Li2018GFRNet}
Xiaoming Li, Ming Liu, Yuting Ye, Wangmeng Zuo, Liang Lin, and Ruigang Yang.
\newblock Learning warped guidance for blind face restoration.
\newblock In {\em ECCV}, 2018.

\bibitem{Lim2017EDSR}
Bee Lim, Sanghyun Son, Heewon Kim, Seungjun Nah, and Kyoung~Mu Lee.
\newblock Enhanced deep residual networks for single image super-resolution.
\newblock In {\em CVPRW}, 2017.

\bibitem{Liu2018Partialconv}
Guilin Liu, Fitsum~A. Reda, Kevin Shih, Ting-Chun Wang, Andrew Tao, and Bryan
  Catanzaro.
\newblock Image inpainting for irregular holes using partial convolutions.
\newblock {\em ArXiv}, 2018.

\bibitem{Liu2019Few}
Ming-Yu Liu, Xun Huang, Arun Mallya, Tero Karras, Timo Aila, Jaakko Lehtinen,
  and Jan Kautz.
\newblock Few-shot unsueprvised image-to-image translation.
\newblock {\em Arxiv}, 2019.

\bibitem{Ma2020FSR}
Cheng Ma, Zhenyu Jiang, Yongming Rao, Jiwen Lu, and Jie Zhou.
\newblock Deep face super-resolution with iterative collaboration between
  attentive recovery and landmark estimation.
\newblock In {\em CVPR}, 2020.

\bibitem{Menon2020PULSE}
Sachit Menon, Alexandru Damian, Shijia Hu, Nikhil Ravi, and Cynthia Rudin.
\newblock Pulse: Self-supervised photo upsampling via latent space exploration
  of generative models.
\newblock In {\em CVPR}, 2020.

\bibitem{Mirza2014cGAN}
Mehdi Mirza and Simon Osindero.
\newblock Conditional generative adversarial nets.
\newblock {\em ArXiv}, 2014.

\bibitem{Nishiyama2009Face}
Masashi Nishiyama, Hidenori Takeshima, Jamie Shotton, Tatsuo Kozakaya, and
  Osamu Yamaguchi.
\newblock Facial deblur inference to improve recognition of blurred faces.
\newblock In {\em CVPR}, 2009.

\bibitem{Park2019SPADE}
Taesung Park, Ming-Yu Liu, Ting-Chun Wang, and Jun-Yan Zhu.
\newblock Semantic image synthesis with spatially-adaptive normalization.
\newblock In {\em CVPR}, 2019.

\bibitem{Richardson2020pSp}
Elad Richardson, Yuval Alaluf, Or Patashnik, Yotam Nitzan, Yaniv Azar, Stav
  Shapiro, and Daniel Cohen-Or.
\newblock Encoding in style: a stylegan encoder for image-to-image translation.
\newblock {\em Arxiv}, 2020.

\bibitem{Ronneberger2015Unet}
Olaf Ronneberger, Philipp Fischer, and Thomas Brox.
\newblock Unet: a convolutional network for biomedical image segmentation.
\newblock {\em Arxiv}, 2015.

\bibitem{Shen2018Deblur}
Ziyi Shen, Wei-sheng Lai, Tingfa Xu, Jan Kautz, and Ming-Hsuan Yang.
\newblock Deep semantic face deblurring.
\newblock In {\em CVPR}, 2018.

\bibitem{Slossberg2018Texture}
Ron Slossberg, Gil Shamai, and Ron Kimmel.
\newblock High quality facial surface and texture synthesis via generative
  adversarial networks.
\newblock In {\em ICCV}, 2018.

\bibitem{Suarez2017Color}
Patricia~L. Suarez, Angel~D. Sappa, and Boris~X. Vintimilla.
\newblock Infrared image colorization based on a triplet dcgan architecture.
\newblock In {\em CVPRW}, 2017.

\bibitem{Wang2018Pix2PixHD}
Ting-Chun Wang, Ming-Yu Liu, Jun-Yan Zhu, Andrew Tao, Jan Kautz, and Bryan
  Catanzaro.
\newblock High-resolution image synthesis and semantic manipulation with
  conditional gans.
\newblock In {\em CVPR}, 2018.

\bibitem{Wang2018ESRGAN}
Xintao Wang, Ke Yu, Shixiang Wu, Jinjin Gu, Yihao Liu, Chao Dong, Yu Qiao, and
  Chen~Change Loy.
\newblock Esrgan: Enhanced super-resolution generative adversarial networks.
\newblock In {\em ECCVW}, 2018.

\bibitem{Wang2020MineGAN}
Yaxing Wang, Abel Gonzalez-Garcia, David Berga, Luis Herranz, Fahad~Shahbaz
  Khan, and Joost van~de Weijer.
\newblock Minegan: effective knowledge transfer from gans to target domains
  with few images.
\newblock In {\em CVPR}, 2020.

\bibitem{Wang2018TransferGAN}
Yaxing Wang, Chenshen Wu, Luis Herranz, Joost van~de Weijer, Abel
  Gonzalez-Garcia, and Bogdan Raducanu.
\newblock Transferring gans: generating images from limited data.
\newblock In {\em ECCV}, 2018.

\bibitem{Yang2020HiFaceGANFR}
Lingbo Yang, Chang Liu, Pan Wang, Shanshe Wang, Peiran Ren, Siweia Ma, and Gao
  Wen.
\newblock Hifacegan: Face renovation via collaborative suppression and
  replenishment.
\newblock {\em Arxiv}, 2020.

\bibitem{Yu2018Deepfill}
Jiahui Yu, Zhe Lin, Jimei Yang, Xiaohui Shen, Xin Lu, and Thomas~S Huang.
\newblock Generative image inpainting with contextual attention.
\newblock {\em ArXiv}, 2018.

\bibitem{Yu2017Hallucinating}
Xin Yu and Fatih Porikli.
\newblock Hallucinating very low-resolution unaligned and noisy face images by
  transformative discriminative autoencoders.
\newblock In {\em CVPR}, 2017.

\bibitem{Zhang2011Sparseprior}
Haichao Zhang, Jianchao Yang, Yanning Zhang, Nasser~M. Nasrabadi, and Thomas~S.
  Huang.
\newblock Close the loop: Joint blind image restoration and recognition with
  sparse representation prior.
\newblock In {\em ICCV}, 2011.

\bibitem{Zhang2018LPIPS}
Richard Zhang, Phillip Isola, Alexei~A Efros, Eli Shechtman, and Oliver Wang.
\newblock The unreasonable effectiveness of deep features as a perceptual
  metric.
\newblock In {\em CVPR}, 2018.

\bibitem{Zhu2017CycleGAN}
Jun-Yan Zhu, Taesung Park, Phillip Isola, and Alexei~A Efros.
\newblock Unpaired image-to-image translation using cycle-consistent
  adversarial networks.
\newblock In {\em ICCV}, 2017.

\bibitem{Zhu2017Multimodal}
Jun-Yan Zhu, Richard Zhang, Deepak Pathak, Trevor Darrell, Alexei~A Efros,
  Oliver Wang, and Eli Shechtman.
\newblock Toward multimodal image-to-image translation.
\newblock In {\em NIPS}, 2017.

\end{thebibliography}
}

\end{document}